\documentclass[letterpaper, 10 pt, conference]{ieeeconf}  

\IEEEoverridecommandlockouts                              

\usepackage{amsmath,amssymb,amsfonts}
\usepackage{algorithmic}
\usepackage{graphicx}
\usepackage{graphics}
\usepackage{textcomp}
\usepackage[table,xcdraw]{xcolor}
\usepackage{leftindex}
\usepackage{amsmath}
\usepackage{mathtools}
\usepackage{multirow}
\usepackage{booktabs}
\usepackage{svg}
\usepackage{hyperref}
\hypersetup{
    colorlinks=true,
    linkcolor=black,
    filecolor=magenta,      
    urlcolor=blue,
    }
\usepackage{calrsfs}
\usepackage{censor}
\usepackage{upgreek}
\usepackage{tikz}
\usetikzlibrary{3d,calc,intersections, positioning, shapes.geometric, fit}
\usepackage{mathrsfs} 
\usepackage{bm}
\usepackage{makecell}

\def\BibTeX{{\rm B\kern-.05em{\sc i\kern-.025em b}\kern-.08em
    T\kern-.1667em\lower.7ex\hbox{E}\kern-.125emX}}

\definecolor{softgray}{gray}{0.6} 

\usepackage{etoolbox}
\makeatletter
\patchcmd{\@makecaption}
  {\scshape}
  {}
  {}
  {}
\makeatother

\newcommand{\topTableSpacing}{2pt}
\newcommand{\bottomTableSpacing}{1pt}

\usepackage{fancyhdr}
\usepackage{xcolor}
\fancypagestyle{arxiv}{%
  \fancyhf{}
  
  \fancyhead[C]{\smash{\raisebox{0.75cm}{\parbox{\textwidth}{\color{darkgray}\scriptsize
    {\centering \textbf{Accepted for publication in the 2026 IEEE/RSJ International Conference on Intelligent Robots and Systems (IROS).}\\[0.5ex]}
    \copyright~2026 IEEE. Personal use of this material is permitted. Permission from IEEE must be obtained for all other uses, in any current or future media, including reprinting/republishing this material for advertising or promotional purposes, creating new collective works, for resale or redistribution to servers or lists, or reuse of any copyrighted component of this work in other works.
  }}}}
}

\title{\LARGE \bf
GLAM-SLAM: Real-time Gaussian Large-scale Mapping via \\ Flow Densification and Spatial Decomposition
}

\author{Panagiotis Mermigkas$^{1, 2, 3}$, Argyris Manetas$^{2}$ and Petros Maragos$^{1, 2, 3}$
\thanks{*\url{https://github.com/pmermigkas/GLAM-SLAM}}
\thanks{$^{1}$HERON - Hellenic Robotics Center of Excellence, Athena Research Center, Greece.}
\thanks{$^{2}$Institute of Robotics, Athena Research Center, Marousi, Greece.}
\thanks{$^{3}$Electrical and Computer Engineering, National Technical University of Athens, 15780 Zografou, Athens, Greece.}
\thanks{{\tt\scriptsize pmermigkas@central.ntua.gr, a.manetas@athenarc.gr, maragos@cs.ntua.gr} \newline
        \hspace{5cm}
        \parbox[c]{0.08\linewidth}{\includegraphics[width=\linewidth]{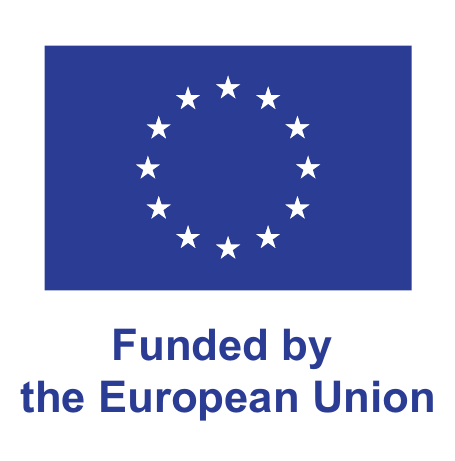}} \hspace{0.01\linewidth} \parbox[c]{0.88\linewidth}{This project is funded by the European Union under Horizon Europe (grant No. 101136568 - HERON).}}
}

\begin{document}

\maketitle
\thispagestyle{arxiv}
\pagestyle{empty}

\begin{abstract}

Existing Gaussian-splatting-based monocular Simultaneous Localization and Mapping (SLAM) systems are either tailored to short sequences, are not real-time, or suffer from prohibitive GPU memory requirements, limiting their applicability in realistic, long-horizon scenarios.
To address this, we present GLAM-SLAM, a real-time, decoupled Gaussian-splatting SLAM system designed for large-scale outdoor scenes.
We ensure lightweight tracking using a robust, feature-based SLAM frontend, while for mapping, we adopt a structured, sparse anchor grid representation that ensures scalable operation and maintains scene coherence across long-term sequences.
To satisfy the dense initialization requirements of 3D Gaussian Splatting (3DGS), we introduce a geometry-based flow-densification anchoring strategy using epipolar constraints. Furthermore, by treating mapping as a multi-scene problem, we propose a scene-partitioning strategy that introduces a strong spatial inductive bias via MLP initializations to generate localized Gaussians.
We evaluate our system on the challenging, long-sequence KITTI Odometry, Oxford RobotCar, and M\'alaga datasets. Extensive ablations and comparisons demonstrate a 15\% improvement in reconstruction quality over the second-best performer, while maintaining real-time performance and the ability to scale to longer sequences. Code is publicly available for the benefit of the community*.

\end{abstract}    
\section{Introduction}

Visual SLAM systems have seen significant improvements in localization accuracy, tracking speed, and robustness, all while maintaining the requirement for real-time execution. Beyond these improvements, the evolution in map representability, from sparse and semi-dense representations~\cite{murartal2017:orbslam2, newcombe2011:kinectfusion}, to NeRF-based~\cite{zhu2022:niceslam, sandstrom2023:pointslam} and eventually 3DGS SLAM systems~\cite{matsuki2024:gaussian, keetha2024:splatam, huang2024:photoslam} led to increasingly higher-quality 3D maps. However, effectively scaling 3DGS for photorealistic reconstruction in long-duration sequences remains significantly underexplored.

Historically, early systems prioritized computational efficiency, accurate ego-motion estimation and robustness through sparse, feature-based representations~\cite{murartal2017:orbslam2}. However, their limitation in tasks requiring dense geometric or semantic understanding spurred the development of semi-dense point-cloud~\cite{engel2014:lsd} and continuous surface~\cite{newcombe2011:kinectfusion} approaches to reconstruct the environment by leveraging direct photometric and dense geometric alignment, respectively. These methods offered a paradigm shift from representing the world as a set of distinct features to modeling it as a visually interpretable entity, enabling robotic downstream tasks, such as navigation, object manipulation and semantic scene understanding.

Despite these advancements in map quality, dense SLAM systems remain notoriously resource-intensive, demanding significant memory and GPU computation even for tracking, which often confines them to powerful hardware. They are also highly susceptible to failure in dynamic environments, where moving objects can introduce severe artifacts and corrupt the map. Furthermore, their reliance on photometric consistency makes them vulnerable to overexposure and motion blur. Consequently, while sparse methods excel in trajectory accuracy and long-term stability, their maps are often unusable for interaction; meanwhile, dense methods offer unparalleled scene representational capacity but at a high operational cost and reduced tracking robustness.

\begin{figure}
    \centering
    \includegraphics[width=\columnwidth]{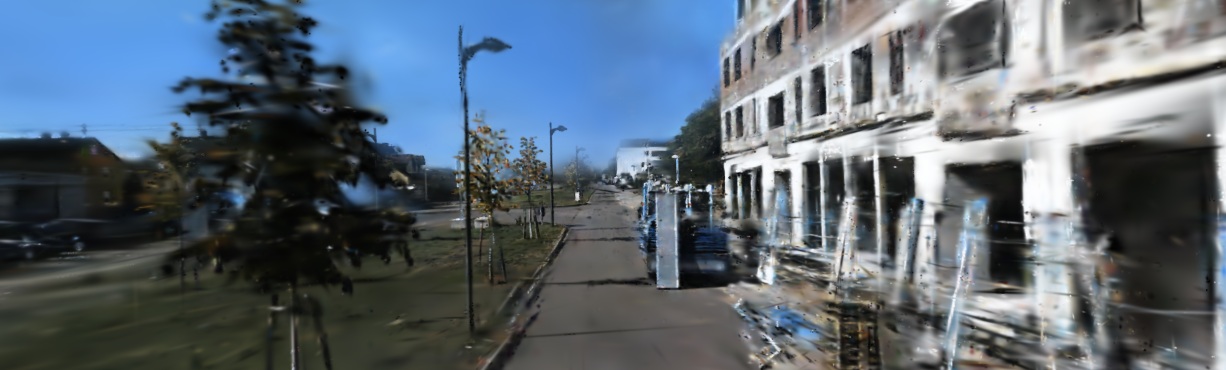}
    
    \vspace{0.05cm}

    \centering
    \includegraphics[width=0.75\columnwidth]{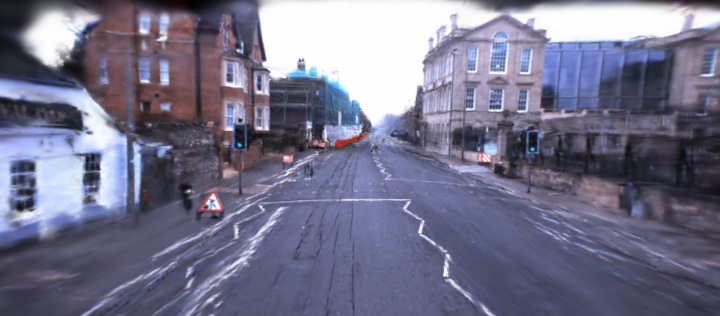}
    
    \vspace{-0.5\baselineskip}
    \caption{\textbf{Novel view synthesis from unseen trajectories:} These renderings, generated from camera poses not present in the KITTI (top) and Oxford RobotCar (bottom) datasets, demonstrate the strong generalization and geometric consistency of our 3DGS-based map without any post-optimization.}
    \vspace{-0.5\baselineskip}
    \label{fig:intro_novelviews}
\end{figure}

To harness the best attributes of both paradigms, our proposed system builds upon the feature-based ORB-SLAM2~\cite{murartal2017:orbslam2} and diverges from traditional mapping by anchoring our parallelized 3DGS reconstruction module upon this robust, sparse scaffolding ---a module that operates on a dedicated GPU to concurrently build a photorealistic, dense map without compromising real-time tracking performance.

However, classic 3DGS implementations present two inherent limitations when applied to long-sequence, sparse, feature-based SLAM. First, the lack of landmark density in sparse tracking creates a geometric density mismatch for 3DGS, which traditionally relies on dense initial point clouds for faster, stable Gaussian optimization, a challenge previously observed in~\cite{RainGS, LiberatedGS}. Second, the prohibitive memory footprint of conventional 3DGS effectively restricts reconstruction to single, isolated environments. While anchor-based approaches like Scaffold-GS~\cite{lu2024:scaffoldgs} attempt to address this by representing multiple Gaussian parameters with a global MLP set, they often struggle with the environmental and illumination variability inherent in expansive, long-sequence data.

In this work, we first propose a Flow-Guided Densification Module that leverages optical flow to recover epipolar-consistent correspondences, thereby addressing initialization bias. This module provides a geometric prior in sparse regions, ensuring uniform coverage of the scene’s spatial manifold. Furthermore, we treat long-sequence reconstruction as a multi-scene challenge and propose a spatial conditioning strategy that partitions the environment and assigns localized MLP sets to estimate Gaussian parameters, substantially improving reconstruction fidelity across varied and large-scale trajectories. Our key contributions are as follows:

\begin{itemize}
    \item A decoupled architecture for real-time photorealistic SLAM with scalable, reduced GPU memory footprint.
    \item A Flow-Guided Densification Module for the initialization of Gaussian anchors and a region-adaptive strategy to capture varying local map conditions.
    \item Detailed ablation studies on our contributions.
    \item Extensive photometric and computational evaluation on challenging, \textit{outdoor, long-sequence} datasets.
\end{itemize}
\section{Related Work}

Generating precise and meaningful scene representations in real time, without compromising accurate camera tracking, remains a central focus in visual SLAM. To surpass the limitations of sparse maps, the field has heavily investigated dense point cloud architectures~\cite{engel2014:lsd, whelan2015:elasticfusion} and volumetric TSDF representations~\cite{newcombe2011:kinectfusion, whelan2012:kintinuous} ---which uniquely allow for the extraction of coherent geometric surfaces. While both paradigms significantly improve overall scene comprehensibility, they inherently struggle to capture fine texture and detail, resulting in reconstructions that lack photorealism.

A step towards photorealistic maps was marked by the pioneering work on Neural Radiance Fields NeRF~\cite{mildenhall2021:nerf}, which demonstrated that Multi-Layer Perceptrons (MLPs) can effectively approximate the radiance field function, achieving photorealistic novel view synthesis without explicit geometry. Due to its representational capabilities, subsequent SLAM systems have adopted this method~\cite{sucar2021:imap, zhu2022:niceslam, sandstrom2023:pointslam, Deng2024:plgslam} to produce online algorithms capable of fine-grained rendering, tracking and photorealistic scene reconstructions. Despite their impressive capabilities to model scenes, these NeRF-based methods are computationally expensive at render time due to the need for a network inference for every unprojected ray and lack editability, since scene modifications require full model retraining.

Recent research has shown growing interest in explicit radiance field representations, most notably 3D Gaussian Splatting in the seminal work of~\cite{kerbl2023:3dgs}. This approach represents a scene with a set of editable 3D Gaussians, each possessing spatial, color and opacity information, and uses a differentiable tile-based rasterizer to achieve real-time rendering and faster training than neural network equivalent methods. This development has sparked a series of works~\cite{chen2024:survey3dgs}; 
2D-GS~\cite{huang2024:2dgs} replaces 3D gaussians with their 2D equivalent primitives to improve surface fidelity, while~\cite{lu2024:scaffoldgs} replaces them with anchor points that spawn multiple Gaussians to reduce memory usage, capture finer details, and accelerate convergence. Specifically, it introduces a sparse voxel grid of anchors, containing descriptors that encode local radiance information (similarly to~\cite{zhu2022:niceslam, sandstrom2023:pointslam}) used to dictate local Gaussians features through global MLPs.
Due to its efficiency, 3D-GS has been adopted in SLAM systems with early \textit{coupled} implementations, which bypass external visual odometry performing tracking using rendering-based optimization directly against the evolving Gaussian map, such as 3DGS-based MonoGS~\cite{matsuki2024:gaussian} and SplaTAM~\cite{keetha2024:splatam} and 2DGS-based method~\cite{zhong2026:globally}.

In contrast, \textit{decoupled} methods maintain reliable tracking performance and mitigate degradation from imperfections in the scene representation by employing an independent tracking module, treating the Gaussian map as a parallel reconstruction backend. These include: (i) neural network–based models for correspondence matching and camera tracking through Dense Bundle Adjustment (DBA) as in VINGS-Mono~\cite{wu2025:vingsmono}, or a Disparity, Scale, and Pose Optimization (DSPO) layer, as in Splat-SLAM~\cite{sandstrom2025:splat_slam}; (ii) ICP-based frontends, as in RTG-SLAM~\cite{peng2024:rtgslam}; (iii) TSDF-fusion tracking frontends, as in GS-Fusion~\cite{wei2024:gsfusion}; and (iv) sparse feature-based SLAM systems, as in the real-time PhotoSLAM~\cite{huang2024:photoslam}. In this work, we adopt the decoupled paradigm, which preserves tracking robustness, preventing mapping errors from degrading localization accuracy, and enhancing the overall modularity and flexibility of our system, in order to handle the demands of long-term operation in unbounded scenes.

Despite significant progress in high-fidelity scene reconstruction, the majority of monocular methods remain confined to constrained indoor environments. Scaling these approaches to unbounded, long-term sequences, such as driving scenarios, remains an underexplored area, as the severe reduction in keyframe covisibility amplifies tracking drift and compromises global map consistency.
For instance, VGGT-Long~\cite{deng2025:vggtlong} addresses SLAM in long outdoor sequences by applying Sim(3) transformations across sliding windows of keyframes to successively merge the camera poses and point clouds generated by a visual foundation model; however, this discrete representation inherently lacks the capacity for photorealistic rendering.
Another approach, GigaSLAM~\cite{deng2025:gigaslam} adapts a Level-of-Detail (LoD) voxel mapping strategy to construct photorealistic representations of extensive scenes. However, its heavy GPU reliance for frontend tasks ---such as feature extraction, matching, depth estimation, and pose estimation--- leaves limited compute time for online Gaussian optimization, forcing a lengthy color refinement phase upon termination that restricts its use in online scenarios.

Our real-time framework leverages a lightweight SLAM frontend to supply high-quality structured anchors to a Gaussian mapping backend, which operates asynchronously on a dedicated GPU.
Moreover, to meet the initialization requirements of 3D-GS~\cite{RainGS, LiberatedGS}, we adopt a geometrically verifiable approach that utilizes optical flow and epipolar constraints to densify the initial point cloud. This provides a highly accurate geometric prior, fundamentally improving upon noisy initialization methods, as in PhotoSLAM~\cite{huang2024:photoslam}, that addresses sparsity by back-projecting untracked features using averaged depths from their neighbors.
Finally, inspired by NeRF partitioning strategies for capturing high-frequency details~\cite{rebain2021:derf} and scaling to large environments~\cite{tancik2022:blocknerf}, we mitigate catastrophic forgetting by dedicating multiple, region-specific MLPs to govern only local Gaussian properties, ensuring that varying outdoor conditions are modeled without overwriting previously mapped regions.
\section{Methodology}
\label{sec:methodology}

\subsection{Preliminaries}

According to the 3DGS representation approach, a scene is modelled as a set of anisotropic 3D Gaussians, with position $\mu_i$, covariance matrix $\Sigma_i$, opacity $\alpha_i$, and view-dependent color expressed through spherical harmonics (SH) coefficients. Using an efficient differentiable rasterization process, these 3D Gaussians are projected onto the 2D image plane, enabling highly parallel rendering and fast photometric loss optimization. Gaussians can be dynamically spawned or pruned during training through an adaptive density mechanism. The rendered color $\mathbf{C}(\mathbf{p})$ at a pixel $\mathbf{p}$ is obtained via alpha compositing over $n$ projected 2D Gaussians:

\begin{equation}
    \mathbf{C}(\mathbf{p}) = \sum_{i=1}^n \mathbf{c}_i \alpha_i(\mathbf{p}) \prod_{j=1}^{i-1} (1 - \alpha_j(\mathbf{p}))
    \label{eq:alpha_blending}
\end{equation}
where $\mathbf{c}_i$ denotes the color of the $i$-th Gaussian, and $\alpha_i(\mathbf{p})$ its opacity at pixel $\mathbf{p}$, evaluated by scaling the projected 2D Gaussian with its corresponding learned opacity.

\textbf{Structured Gaussian Scene Representation:} Our Gaussian mapping module adopts the sparse-based approach introduced in Scaffold-GS~\cite{lu2024:scaffoldgs}, which represents the scene as an anchor grid, keeping memory requirements low and preserving the underlying geometric integrity. Each anchor $a$ encodes local information, comprising a learnable feature descriptor $f_a \in \mathbb{R}^{8}$, a scaling parameter $l_y$, and $k$ trainable offset vectors, used to generate its associated set of $k$ Gaussians. All Gaussian attributes are predicted implicitly by a set of lightweight Multi-Layer Perceptrons (MLPs), conditioned on the anchor feature descriptor, the anchor-to-camera distance $\delta_{ac}$, and the viewing-direction vector $\textbf{d}_{ac}$. For a camera located at $\textbf{x}_c$ and an anchor at $\textbf{x}_a$, these quantities are defined as:
\begin{equation}
    \delta_{ac} = ||\textbf{x}_a - \textbf{x}_c||_2 ,
    \quad
    \mathbf{d}_{ac}=\frac{\textbf{x}_a-\textbf{x}_c}{||\textbf{x}_a-\textbf{x}_c||_2} 
\end{equation}
Training is guided by a combined photometric and structural similarity objective as defined in~\cite{kerbl2023:3dgs}: 
\begin{equation}
    \label{eq:loss}
    \mathcal{L} = \lambda \mathcal{L}_1 + (1-\lambda) \mathcal{L}_{SSIM}
\end{equation}
which balances per-pixel accuracy with structural coherence.

\begin{figure}
    \centering
    \includegraphics[width=0.49\textwidth]{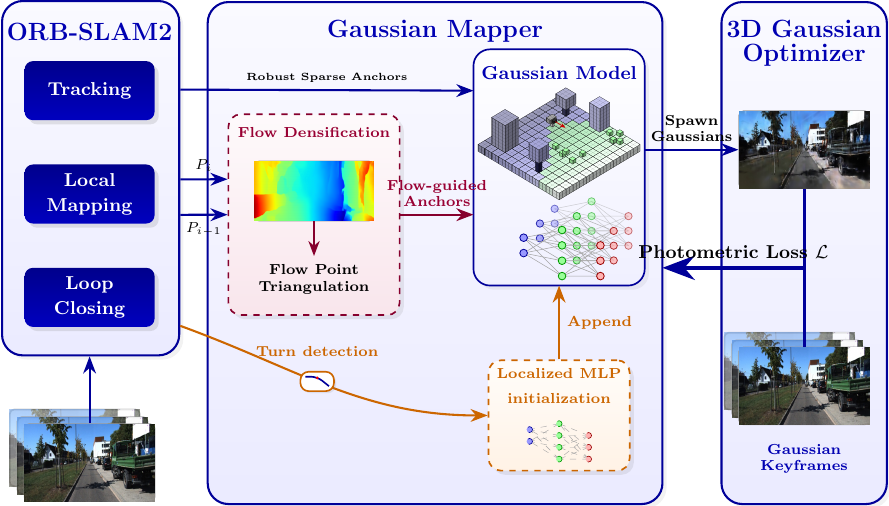}

    \vspace{-0.5\baselineskip}
    \caption{\textbf{GLAM (Gaussian LArge-scale Mapping)-SLAM system overview}: Our system follows a decoupled design with an ORB-SLAM2 thread serving as the real-time front-end and a Gaussian Mapper thread that constructs its anchor grid by integrating sparse ORB-SLAM2 points with flow-densified points.}
    \vspace{-0.5\baselineskip}
    \label{fig:methods_systemoverview}
\end{figure}

\begin{figure*}[ht]
    \centering
    \begin{center}
    \begin{tabular}{@{} c c c}
        \includegraphics[width=0.31\textwidth]{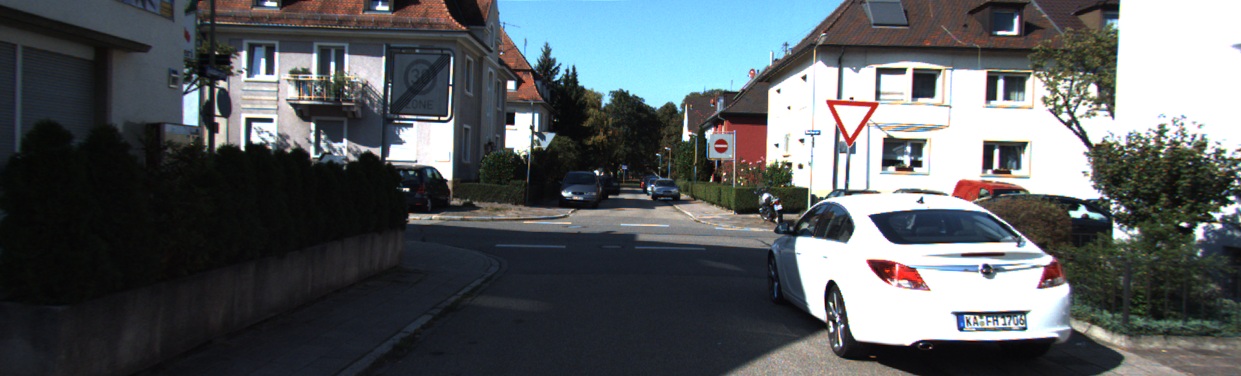} & \includegraphics[width=0.31\textwidth]{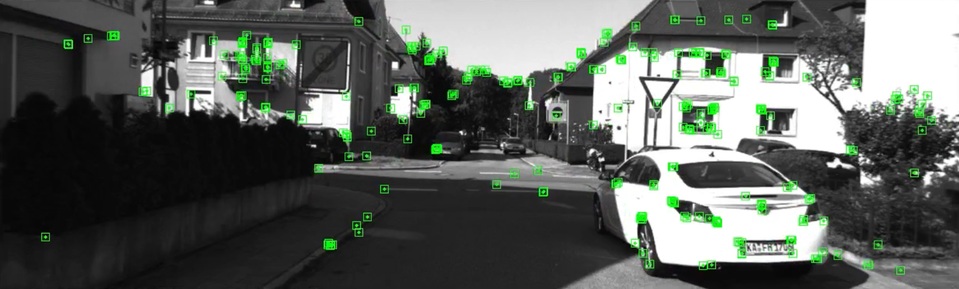} &
        \includegraphics[width=0.31\textwidth]{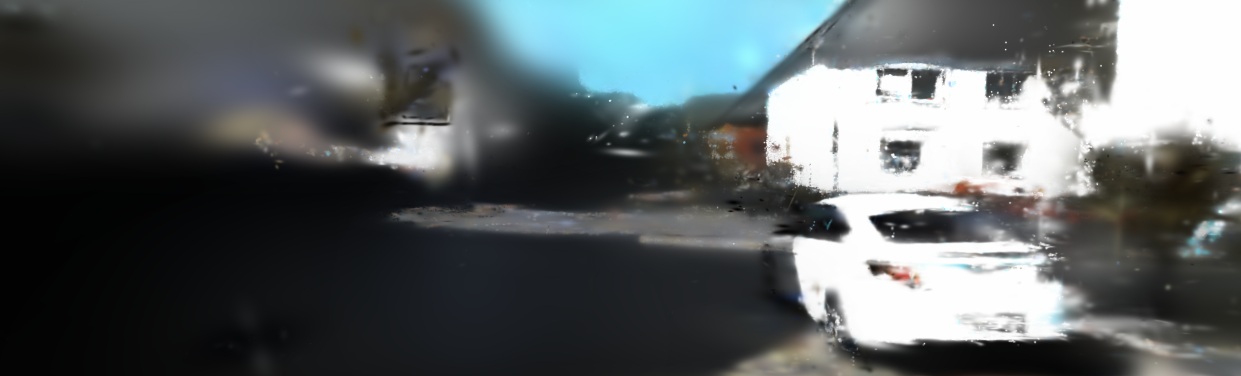} \\
        \scriptsize{(a) Ground Truth image} & \scriptsize{(b) ORB-SLAM2 tracked features} & \scriptsize{(c) Rendered image (Baseline), PSNR = 15.80} \\ [0.1cm] 
        \includegraphics[width=0.31\textwidth]{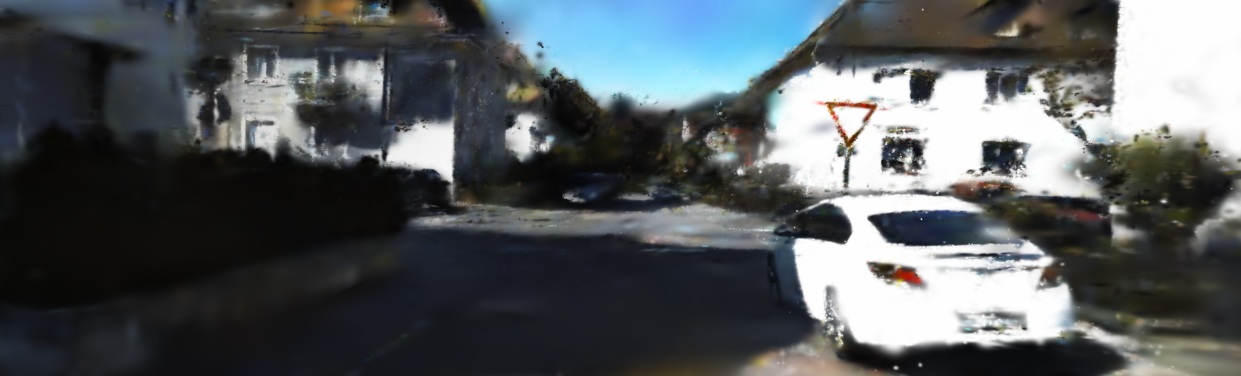} & \includegraphics[width=0.31\textwidth]{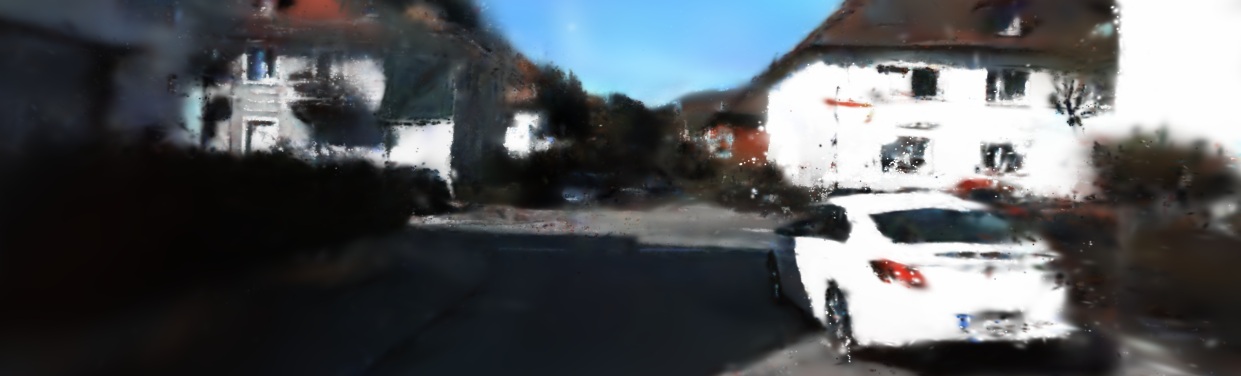} & \includegraphics[width=0.31\textwidth]{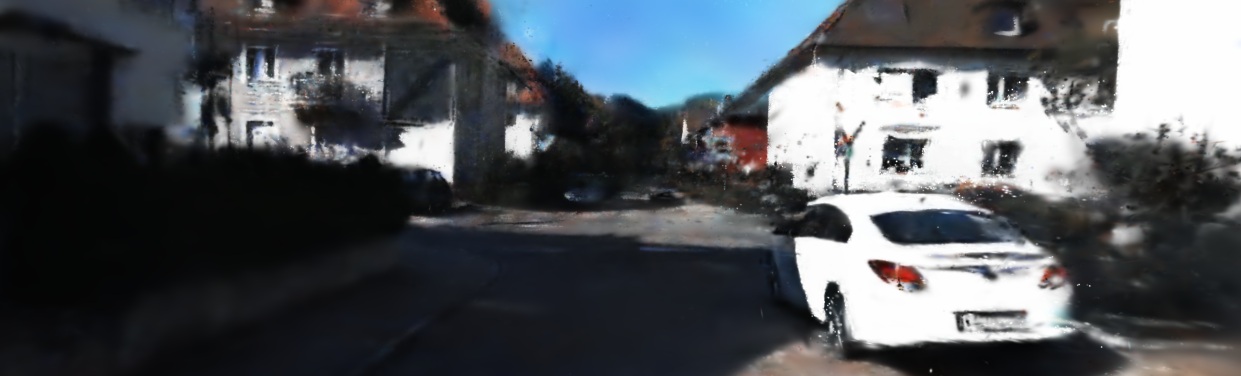} \\
        \scriptsize{(d) Rendered image (OF), PSNR = 17.64} & \scriptsize{(e) Rendered image (MLP), PSNR = 17.37} & \scriptsize{(f) Rendered image (OF + MLP), PSNR = 18.37} \\ [0.1cm]
        
    \end{tabular}
    \end{center}

    \vspace{-0.85\baselineskip}
    \caption{\textbf{Qualitative rendering comparison of our contributions on KITTI Odometry Seq.~\texttt{00}:} The lack of structured anchors derived from ORB-SLAM2 features in the left side of the scene leads to blurry localized renderings in our baseline implementation (c). Visual fidelity in this area is greatly improved when employing our individual contributions (d, e) and their combination (f).}
    \vspace{-0.5\baselineskip}
    \label{fig:supplementary_kitti00_qualitative_295}

\end{figure*}

\subsection{System Overview}
Our system overview is illustrated in Fig.~\ref{fig:methods_systemoverview}. We adopt a decoupled design where the GPU-based Gaussian mapping module operates independently of the tracking subsystem, which runs exclusively on CPU. The system relies on the multi-threaded ORB-SLAM2 framework, whose tracking, local mapping, and loop closing threads run in parallel. Concurrently, the Gaussian Mapping thread maintains the voxelized Gaussian Model and expands it using an incremental variant of Scaffold-GS~\cite{lu2024:scaffoldgs}. This expansion is driven by the integration of ORB-SLAM2 tracked points $\mathbf{P_1} \in \mathbb{R}^{M_1\times3}$ at every keyframe, supplemented by periodic geometric correspondences $\mathbf{P_2} \in \mathbb{R}^{M_2\times3}$ extracted via a lightweight optical flow model. The fused point cloud is discretized onto a voxel grid with centers $\mathbf{V} \in \mathbb{R}^{N\times3}$, referred to as anchors, via the equation:

\begin{equation}
\label{eq:voxelization}
\mathbf{V} = \left\{ \left\lfloor \frac{\mathbf{P_1 \cup P_2}}{\epsilon} \right\rceil \right\} \cdot \epsilon,
\end{equation}
where $\{\cdot \}$ denotes the operation that retains only unique entries, and $\epsilon$ specifies the voxel size.
Furthermore, following substantial rotation changes, new MLPs are introduced to predict the parameters of the spawned Gaussians until another major viewpoint change occurs. The local bundle adjustment module of ORB-SLAM2 continuously refines keyframe poses, resulting in corresponding updates to the poses stored within the Gaussian map. In contrast, once initialized, anchor points are not directly modified by the local operations of the tracker and are adjusted exclusively during loop closures to correct accumulated global drift. The Gaussian optimization process runs independently, performing refinement, densification, and pruning to minimize~\eqref{eq:loss}.

\subsection{Flow Densification}
\label{sec:methodology_opticalflow}

Sparse initializations in 3DGS severely compromise both visual fidelity and the rapid convergence speeds necessary for real-time execution, as evidenced by recent literature~\cite{RainGS, LiberatedGS}. Indeed, ORB-SLAM2, which is used as our Gaussian Mapper anchor initializer, despite providing a strong structure, only retains the most reliable triangulations ---those yielding minimal 2D reprojection errors. This selective process frequently results in an uneven spatial distribution of inlier features with some regions overrepresented and others insufficiently covered. Because the 3DGS optimization struggles to converge in these sparse areas, they exhibit reduced geometric and photometric fidelity, appearing blurred in rendered outputs and yielding lower PSNR values (Fig.~3).

\textbf{Dense Flow.}
To mitigate these issues, we augment the Gaussian map points with geometrically consistent matches obtained using a dense optical flow network~\cite{hui2020liteflownet3}. This network is employed to establish dense correspondences $\mathbf{x} \leftrightarrow \mathbf{x}'$ (between consecutive keyframes), expressed in homogeneous coordinates as $\mathbf{x}' = \mathbf{x} + \mathbf{u}, \qquad \mathbf{x} = [x, y, 1]^\top, \quad \mathbf{u} = [u_x, u_y, 0]^\top$ where \(u_x\) and \(u_y\) represent the horizontal and vertical components of the dense flow vector, respectively.

Many existing frameworks rely on optical flow for per-frame tracking~\cite{sandstrom2025:splat_slam, zhu2025:vigsslam} which violates strict real-time constraints. Instead, to overcome this critical bottleneck we completely decouple tracking from optical flow, reserving optional flow extraction solely for Gaussian mapping. We apply this optical flow sparsely to prevent diminishing returns, as continuous execution reduces total Gaussian optimization iterations and redundantly collapses new points into already-populated anchors. Beyond computational efficiency, this tracking-optical flow decoupling prevents the severe zero-shot degradation caused by the domain bias of deep flow networks, even if they occasionally improve specific in-domain trajectories (e.g. KITTI). This architectural separation is highly effective because tracking demands robust, precise correspondences, whereas mapping can successfully leverage slightly noisy flow-based correspondences to populate sparsely initialized areas.

\textbf{Correspondence Filtering.}
Let \( \mathrm{P}_i = \mathrm{K} [ \mathrm{R}_i \,|\, \mathbf{t}_i ] \) denote the calibrated projection matrix at keyframe \( i \), where the camera poses \((\mathrm{R}_i, \mathbf{t}_i)\) are provided by ORB-SLAM2. For computational efficiency, the epipolar constraint residual is evaluated over \( N \) randomly sampled homogeneous correspondence pairs \(\mathbf{x} \leftrightarrow \mathbf{x}'\):
\begin{equation}
\text{residual} = \left| \mathbf{x}^{'\top} \mathrm{F} \mathbf{x} \right| < \tau
\label{eq:methods_epipolar_residual}
\end{equation}
where the fundamental matrix \( \mathrm{F} = [\mathbf{e}_2]_{\times} \mathrm{P}_2 \mathrm{P}_1^{+} \) is computed using the known epipole in the second image, \(\mathbf{e}_2 = \mathrm{P}_2 \mathbf{C}_1\), with \(\mathbf{C}_1\) being the camera center of the first view (the right null space of \( \mathrm{P}_1 \)). Here, \( \mathrm{P}_1^{+} \) denotes the pseudo-inverse of \( \mathrm{P}_1 \), and \([\cdot]_{\times}\) represents the skew-symmetric matrix operator. Correspondences satisfying \(\text{residual} < \tau\) are retained as robust matches and subsequently triangulated via known camera poses to yield supplementary geometric anchors for the Gaussian map.

\begin{table*}
\setlength{\tabcolsep}{5pt}

\newlength\wexp
\settowidth{\wexp}{Static Line Trackletsssssssssssss}
\newcolumntype{C}{>{\raggedleft\arraybackslash}p{\dimexpr(\wexp-2\tabcolsep)/2}}

\newlength\xexp
\settowidth{\xexp}{1.020}
\newcolumntype{D}{>{\raggedleft\arraybackslash}p{\dimexpr(\xexp)}}

\newlength\yexp
\settowidth{\yexp}{11.161}
\newcolumntype{E}{>{\raggedleft\arraybackslash}p{\dimexpr(\yexp)}}

\newlength\zexp
\settowidth{\zexp}{11.161$^{\times}$}
\newcolumntype{F}{>{\raggedleft\arraybackslash}p{\dimexpr(\zexp)}}

\centering
\caption{Photometric Reconstruction Quality metrics (PSNR, SSIM, LPIPS) on the KITTI Odometry Dataset. $^\times$: Due to out of memory limitations, PhotoSLAM has its reported results for 2000 frames and GigaSLAM for 3200 frames.}
\vspace{-0.75\baselineskip}
\scriptsize\resizebox{0.9\textwidth}{!}{%
\begin{tabular}{@{} l @{}p{0.15cm} C || F E F E E F E E F E E || E }
\toprule[1.5pt] \addlinespace[\topTableSpacing]
\textbf{Method} & & \multicolumn{1}{c||}{\textbf{Sequences} $\rightarrow$} & \multicolumn{1}{c}{\texttt{00}} & \multicolumn{1}{c}{\texttt{01}} & \multicolumn{1}{c}{\texttt{02}} &  \multicolumn{1}{c}{\texttt{03}} & \multicolumn{1}{c}{\texttt{04}} & \multicolumn{1}{c}{\texttt{05}} & \multicolumn{1}{c}{\texttt{06}} & \multicolumn{1}{c}{\texttt{07}} & \multicolumn{1}{c}{\texttt{08}} & \multicolumn{1}{c}{\texttt{09}} & \multicolumn{1}{c||}{\texttt{10}} & \multicolumn{1}{c}{\textbf{Avg.}} \\
  
\textit{Total frames} &  &  & \multicolumn{1}{c}{\textit{4541}} & \multicolumn{1}{c}{\textit{1101}} & \multicolumn{1}{c}{\textit{4661}} & \multicolumn{1}{c}{\textit{801}} & \multicolumn{1}{c}{\textit{271}} & \multicolumn{1}{c}{\textit{2761}} & \multicolumn{1}{c}{\textit{1101}} & \multicolumn{1}{c}{\textit{1101}} & \multicolumn{1}{c}{\textit{4071}} & \multicolumn{1}{c}{\textit{1591}} & \multicolumn{1}{c||}{\textit{1201}} \\ \addlinespace[\topTableSpacing] \hline \addlinespace[\topTableSpacing]

  & & PSNR $\uparrow$\phantom{SSS}                 & \textcolor{softgray}{\textit{11.161}$^{\times}$}                  & 18.336\phantom{.}          & \textcolor{softgray}{\textit{13.613}$^{\times}$}                  & 15.604\phantom{.}          & 15.660\phantom{.}          & \textcolor{softgray}{\textit{12.661}$^{\times}$}          &  12.169\phantom{.}        & 9.497\phantom{.}  & \textcolor{softgray}{\textit{13.374}$^{\times}$} & 10.139\phantom{.} &  15.003\phantom{.} & 13.383\phantom{.}            \\

PhotoSLAM & & SSIM $\uparrow$\phantom{SSS}                & \textcolor{softgray}{\textit{0.433}$^{\times}$}                 & 0.624\phantom{.}          & \textcolor{softgray}{\textit{0.420}$^{\times}$}                  & 0.453\phantom{.}          & 0.516\phantom{.}          & \textcolor{softgray}{\textit{0.447}$^{\times}$} & 0.438\phantom{.}         & 0.376\phantom{.}  & \textcolor{softgray}{\textit{0.446}$^{\times}$} & 0.358\phantom{.}         &  0.490\phantom{.} & 0.455\phantom{.} \\

 & & LPIPS $\downarrow$\phantom{SSS}                &         \textcolor{softgray}{\textit{0.821}$^{\times}$}          &      0.702\phantom{.}     &       \textcolor{softgray}{\textit{0.816}$^{\times}$}          &   0.777\phantom{.}        &    0.792\phantom{.}       &    \textcolor{softgray}{\textit{0.819}$^{\times}$}      &     0.825\phantom{.}      & 0.826\phantom{.} &  \textcolor{softgray}{\textit{0.809}$^{\times}$} &     0.853\phantom{.}      & 0.781\phantom{.} &   0.802\phantom{.}  \\ \addlinespace[\topTableSpacing] \hline \addlinespace[\topTableSpacing]

  & & PSNR $\uparrow$\phantom{SSS}                 &    \textcolor{softgray}{\textit{15.674}$^{\times}$} & 15.647\phantom{.}          & \textcolor{softgray}{\textit{15.542}$^{\times}$}                  & 16.258\phantom{.}          & 15.854\phantom{.}         & 14.736\phantom{$^{\times}$} & 14.983\phantom{.} & \textbf{14.583}\phantom{.}  & \textcolor{softgray}{\textit{16.071}$^{\times}$} & 15.569\phantom{.} & 15.665\phantom{.} & 15.507\phantom{.}              \\

GigaSLAM & & SSIM $\uparrow$\phantom{SSS}                &  \textcolor{softgray}{\textit{0.808}$^{\times}$} & 0.801\phantom{.}          & \textcolor{softgray}{\textit{0.777}$^{\times}$}                  & 0.763\phantom{.}          & 0.709\phantom{.}         & 0.758\phantom{$^{\times}$} & 0.785\phantom{.} & 0.712\phantom{.}            & \textcolor{softgray}{\textit{0.801}$^{\times}$} & 0.751\phantom{.}         & 0.736\phantom{.} & 0.764\phantom{.} \\

(pre-opt) & & LPIPS $\downarrow$\phantom{SSS}                & \textcolor{softgray}{\textit{0.786}$^{\times}$} &  0.770\phantom{.}         & \textcolor{softgray}{\textit{0.781}$^{\times}$}                 & 0.818\phantom{.}          & 0.718\phantom{.}         & 0.833\phantom{$^{\times}$} & 0.790\phantom{.}         & 0.779\phantom{.}         & \textcolor{softgray}{\textit{0.786}$^{\times}$} & 0.793\phantom{.}           & 0.808\phantom{.} & 0.787\phantom{.}    \\ \addlinespace[\topTableSpacing] \hline \addlinespace[\topTableSpacing]

& & PSNR $\uparrow$\phantom{SSS} &  & 13.914\phantom{.} &  &  & 14.670\phantom{.} &  & 8.141\phantom{.} &  &  &  &  &        \multicolumn{1}{c}{\phantom{s}-\phantom{.}}        \\

MonoGS & & SSIM $\uparrow$\phantom{SSS} & \textcolor{softgray}{\textit{failed}}\phantom{$^{\times}$} & 0.523\phantom{.} & \textcolor{softgray}{\textit{failed}}\phantom{$^{\times}$} & \textcolor{softgray}{\textit{failed}}\phantom{.} & 0.450\phantom{.} & \textcolor{softgray}{\textit{failed}}\phantom{$^{\times}$} & 0.184\phantom{.} & \textcolor{softgray}{\textit{failed}}\phantom{.} & \textcolor{softgray}{\textit{failed}}\phantom{$^{\times}$} & \textcolor{softgray}{\textit{failed}}\phantom{.} & \textcolor{softgray}{\textit{failed}}\phantom{.} & \multicolumn{1}{c}{\phantom{s}-\phantom{.}} \\

 & & LPIPS $\downarrow$\phantom{SSS} &  & 0.702\phantom{.} &  &  & 0.691\phantom{.} &  & 0.918\phantom{.} &  &  &  &  &  \multicolumn{1}{c}{\phantom{s}-\phantom{.}}  \\ \addlinespace[\topTableSpacing] \hline \addlinespace[\topTableSpacing]

  & & PSNR $\uparrow$\phantom{SSS} & 15.096\phantom{$^{\times}$} & \textbf{20.152}\phantom{.} & 13.351\phantom{$^{\times}$} & \textbf{20.254}\phantom{.} & \textbf{21.509}\phantom{.} & \textbf{14.889}\phantom{$^{\times}$} & \textbf{16.405}\phantom{.} & 13.624\phantom{.} & \textbf{17.158}\phantom{$^{\times}$} & \textbf{19.185}\phantom{.} & \textbf{18.754}\phantom{.} & \textbf{17.307}\phantom{.}              \\

Ours & & SSIM $\uparrow$\phantom{SSS} & 0.802\phantom{$^{\times}$} & \textbf{0.909}\phantom{.} & 0.715\phantom{$^{\times}$} & \textbf{0.889}\phantom{.} & \textbf{0.922}\phantom{.} & \textbf{0.789}\phantom{$^{\times}$} & \textbf{0.831}\phantom{.} & \textbf{0.721}\phantom{.} & \textbf{0.831}\phantom{$^{\times}$} & \textbf{0.885}\phantom{.} & \textbf{0.875}\phantom{.} & \textbf{0.834}\phantom{.} \\

 & & LPIPS $\downarrow$\phantom{SSS} & \textbf{0.725}\phantom{$^{\times}$} & \textbf{0.552}\phantom{.} & \textbf{0.770}\phantom{$^{\times}$} & \textbf{0.404}\phantom{.} & \textbf{0.416}\phantom{.} & \textbf{0.632}\phantom{$^{\times}$} & \textbf{0.601}\phantom{.} & \textbf{0.590}\phantom{.} & \textbf{0.556}\phantom{$^{\times}$} & \textbf{0.482}\phantom{.} & \textbf{0.512}\phantom{.} & \textbf{0.567}\phantom{.}   \\ \addlinespace[\topTableSpacing] \hline \hline \addlinespace[\topTableSpacing]

  & & PSNR $\uparrow$\phantom{SSS}                 & 9.768\phantom{$^{\times}$}                  & 25.909\phantom{.}          & 21.393\phantom{$^{\times}$} &  24.109\phantom{.}         & 25.188\phantom{.}          &     22.114\phantom{$^{\times}$}      & 23.874\phantom{.}         & 23.904\phantom{.}  & 22.706\phantom{$^{\times}$} & 22.924\phantom{.}  & 23.357\phantom{.} &    22.295\phantom{.}           \\

GigaSLAM & & SSIM $\uparrow$\phantom{SSS}                &  0.583\phantom{$^{\times}$}                & 0.967\phantom{.}          & 0.930\phantom{$^{\times}$} &  0.949\phantom{.}         & 0.959\phantom{.}          & 0.934\phantom{$^{\times}$} & 0.954\phantom{.}         & 0.957\phantom{.}  & 0.940\phantom{$^{\times}$} & 0.941\phantom{.}         & 0.948\phantom{.} & 0.915\phantom{.} \\

(post-opt) & & LPIPS $\downarrow$\phantom{SSS}                &  0.816\phantom{$^{\times}$}                 &  0.336\phantom{.} & 0.412\phantom{$^{\times}$} &  0.367\phantom{.}         & 0.305\phantom{.}          &     0.404\phantom{$^{\times}$}     & 0.358\phantom{.}          & 0.304\phantom{.} & 0.367\phantom{$^{\times}$} & 0.367\phantom{.}          & 0.364\phantom{.} &   0.400\phantom{.}  \\ \addlinespace[\bottomTableSpacing] \bottomrule[1.5pt]

\end{tabular}
}
\vspace{-\baselineskip}
\label{table:results_qualitymetricsmain}

\end{table*}

\subsection{Localized MLP Initialization}
\label{sec:methodology_regionadaptive}

Typically, Gaussian Splatting SLAM method evaluations are confined to bounded scenes (like indoor rooms, vehicles and plants) and contained outdoor areas (such as gardens or distant views of buildings), where a single network provides an adequate and computationally efficient model of the entire space~\cite{lu2024:scaffoldgs, wen2025:segs_slam}. However, in large-scale outdoor datasets, illumination can vary significantly, objects may appear at widely different distances and scales, and surface properties such as reflectivity and color can change across the scene. Under these conditions, a single MLP, especially in real-time SLAM, struggles to capture the full range of spatial and photometric variability, leading to loss of detail and reduced representational fidelity.

To address this, we partition the mapped environment into a dynamic set of $M$ distinct regions, denoted as $\mathcal{R} = \{R_1, R_2, \dots, R_M\}$, with a respective set of independent parameter sets $\Theta = \{\theta_1, \theta_2, \dots, \theta_M\}$. Consequently, the spatial domain $R_m$, to which $\mathbf{C}$ belongs, explicitly conditions the attribute decoding for all $k$ Gaussians:
\begin{equation}
    \mathrm{G} = \sum_{m=1}^{M} \mathbf{1}(\mathbf{C} \in R_m) \cdot \Phi_{\theta_m}(f_a,\delta_{a c}, \mathbf{d}_{a c})
\end{equation}
spawned from an observed anchor $a$, where the indicator function $\mathbf{1}(\cdot)$ activates exclusively for the local MLP assigned to region $R_m$.
Since our spatial decomposition framework is agnostic to the specific initialization trigger, we employ turn detection as an efficient heuristic to demonstrate its efficacy.
Therefore, our real-time system operates entirely without a priori knowledge, bypassing the limitations of~\cite{tancik2022:blocknerf}, which strictly depends on predetermined regions, and~\cite{rebain2021:derf}, which requires knowing the total number of regions beforehand. The results section includes an ablation study that evaluates the effect of incorporating this modification.

\subsection{Loop Closures} 

When the tracking module of ORB-SLAM2 detects a loop closure, corrections are propagated to the Gaussian map by refining the affected keyframe poses. Specifically, for each keyframe $i$ whose pose is updated to $\prescript{c}{w}{\mathrm{T}}_{\text{new}, i}$, we compute the relative transformation:
\begin{equation}    
    \Delta \mathrm{T}_i = (\prescript{c}{w}{\mathrm{T}}_{\text{new}, i})^{-1} \prescript{c}{w}{\mathrm{T}}_{\text{old}, i}
\end{equation}
Each transformation $\Delta \mathrm{T}_i$ is then applied to all anchors within that keyframe's local view frustum, resulting in a non-rigid transformation of the grid. This deformation is essential for maintaining reconstruction consistency and constraining long-term drift.
\section{Results}

\subsection{Experimental Setup}

We evaluate our framework on large-scale, real-world, outdoor driving sequences from the KITTI Odometry~\cite{geiger2013:kitti}, Oxford RobotCar~\cite{maddern2017:RobotCarDataset}, and M\'alaga~\cite{blancoclaraco2014:malaga} datasets. Their diverse environments ---spanning urban streets, highways, and rural roads--- provide a comprehensive benchmark for validating the scalability, robustness and accuracy of the 3DGS reconstructions. 
Our system is compared against three state-of-the-art Gaussian Splatting SLAM approaches on an Intel Core Ultra 9 285K with NVIDIA RTX 5090 (32 GB of VRAM): PhotoSLAM~\cite{huang2024:photoslam}, a real-time, high-quality reconstruction system evaluated primarily for bounded scenes; GigaSLAM~\cite{deng2025:gigaslam}, a long-trajectory system evaluated both with and without its expensive post-processing refinement for fairness; and MonoGS~\cite{matsuki2024:gaussian}, a seminal Gaussian Splatting SLAM baseline. 
Rendering quality is assessed using Peak Signal-to-Noise Ratio (PSNR), Structural Similarity Index Measure (SSIM), and Learned Perceptual Image Patch Similarity (LPIPS), and system performance using FPS, peak GPU memory usage and the number of Gaussian primitives.

\begin{table}[t]
\setlength{\tabcolsep}{5pt}

\settowidth{\wexp}{Static Line Trackletsssssssssssss}
\newcolumntype{C}{>{\raggedleft\arraybackslash}p{\dimexpr(\wexp-2\tabcolsep)/2}}

\settowidth{\xexp}{15.020}
\newcolumntype{D}{>{\raggedleft\arraybackslash}p{\dimexpr(\xexp)}}

\settowidth{\yexp}{15.40000}
\newcolumntype{E}{>{\raggedleft\arraybackslash}p{\dimexpr(\yexp)}}

\centering
\caption{Photometric Reconstruction Quality metrics (PSNR, SSIM, LPIPS) on the Oxford (left) and Malaga (right) Datasets.}
\vspace{-0.75\baselineskip}
\resizebox{0.49\textwidth}{!}{%
\begin{tabular}{@{} l @{}p{0.15cm} C || *{3}{E} || D || *{3}{E} || D }
\toprule[1.5pt] \addlinespace[\topTableSpacing]
\textbf{Method} & & \multicolumn{1}{c||}{\textbf{Sequences} $\rightarrow$} & \multicolumn{1}{>{\centering\arraybackslash}p{\yexp}}{\tiny\texttt{2014-05-14 13-53-47}} & \multicolumn{1}{>{\centering\arraybackslash}p{\yexp}}{\tiny\texttt{2014-08-11 10-22-21}} & \multicolumn{1}{>{\centering\arraybackslash}p{\yexp}||}{\tiny\texttt{2015-11-02 10-45-59}} & \multicolumn{1}{c||}{\textbf{Avg.}} & \multicolumn{1}{c}{\texttt{02}} & \multicolumn{1}{c}{\texttt{06}} & \multicolumn{1}{c||}{\texttt{08}} & \multicolumn{1}{c}{\textbf{Avg.}} \\
  
\textit{Total frames} &  &  & \multicolumn{1}{c}{3032} & \multicolumn{1}{c}{3003} & \multicolumn{1}{c||}{3601} &  & \multicolumn{1}{c}{1855} & \multicolumn{1}{c}{1118} & \multicolumn{1}{c||}{1201} &  \\ \addlinespace[\topTableSpacing] \hline \addlinespace[\topTableSpacing]

  & & PSNR $\uparrow$\phantom{Ssi}   & 13.749\phantom{6} & 14.087\phantom{6} & 11.068\phantom{6} & 12.968 & 17.440\phantom{6} & 21.106\phantom{6} & 19.968\phantom{6} & 19.504 \\
PhotoSLAM & & SSIM $\uparrow$\phantom{Ssi} & 0.529\phantom{6} & 0.554\phantom{6} & 0.311\phantom{6} & 0.465 & 0.667\phantom{6} & 0.740\phantom{6} & 0.705\phantom{6} & 0.704 \\
 & & LPIPS $\downarrow$\phantom{Ssi} & 0.817\phantom{6} & 0.802\phantom{6} & 0.849\phantom{6} & 0.823 & 0.735\phantom{6} & 0.699\phantom{6} & 0.713\phantom{6} & 0.716 \\ \addlinespace[\topTableSpacing] \hline \addlinespace[\topTableSpacing]

  & & PSNR $\uparrow$\phantom{Ssi}   & 14.880\phantom{6} & 15.958\phantom{6} & 16.064\phantom{6} & 15.634 & 19.257\phantom{6} & 20.684\phantom{6} & 19.979\phantom{6} & 19.973 \\
  
GigaSLAM & & SSIM $\uparrow$\phantom{Ssi}  & 0.769\phantom{6} & 0.734\phantom{6} & 0.853\phantom{6} & 0.785 & 0.926\phantom{6} & 0.868\phantom{6} & 0.882\phantom{6} & 0.892 \\

(pre-opt) & & LPIPS $\downarrow$\phantom{Ssi}  & 0.644\phantom{6} & 0.652\phantom{6} & 0.698\phantom{6} & 0.665 & 0.666\phantom{6} & 0.639\phantom{6} & 0.667\phantom{6} & 0.657 \\ \addlinespace[\topTableSpacing] \hline \addlinespace[\topTableSpacing]

  & & PSNR $\uparrow$\phantom{Ssi} & 7.009\phantom{6} & 10.194\phantom{6} & 7.145\phantom{6} & 8.116 & 15.424\phantom{6} & 17.447\phantom{6} & 16.986\phantom{6} & 16.619 \\
  
MonoGS & & SSIM $\uparrow$\phantom{Ssi} & 0.198\phantom{6} & 0.410\phantom{6} & 0.343\phantom{6} & 0.317 & 0.634\phantom{6} & 0.672\phantom{6} & 0.650\phantom{6} & 0.652 \\

 & & LPIPS $\downarrow$\phantom{Ssi} & 0.875\phantom{6} & 0.788\phantom{6} & 0.864\phantom{6} & 0.842 & 0.711\phantom{6} & 0.741\phantom{6} & 0.750\phantom{6} & 0.734 \\ \addlinespace[\topTableSpacing] \hline \addlinespace[\topTableSpacing]

  & & PSNR $\uparrow$\phantom{Ssi}   & \textbf{17.838}\phantom{6} & \textbf{20.827}\phantom{6}  & \textbf{20.719}\phantom{6} & \textbf{19.795} & \textbf{22.332}\phantom{6} & \textbf{23.812}\phantom{6} & \textbf{23.180}\phantom{6} & \textbf{23.108} \\
  
Ours & & SSIM $\uparrow$\phantom{Ssi}  & \textbf{0.883}\phantom{6} & \textbf{0.903}\phantom{6} & \textbf{0.938}\phantom{6} & \textbf{0.908} & \textbf{0.961}\phantom{6} & \textbf{0.942}\phantom{6} & \textbf{0.941}\phantom{6} & \textbf{0.948} \\

 & & LPIPS $\downarrow$\phantom{Ssi} & \textbf{0.527}\phantom{6} & \textbf{0.382}\phantom{6} & \textbf{0.382}\phantom{6} & \textbf{0.430} & \textbf{0.533}\phantom{6} & \textbf{0.492}\phantom{6} & \textbf{0.509}\phantom{6} & \textbf{0.511} \\ \addlinespace[\bottomTableSpacing] \bottomrule[1.5pt]

\end{tabular}
}
\vspace{-0.6\baselineskip}
\label{table:results_qualitymetricsextra}
\end{table}

\subsection{Implementation Details}

The anchor voxel size was set to the default size of 0.001~\cite{lu2024:scaffoldgs} across all sequences; larger sizes cause multiple initialization points to collapse into a single anchor, which severely diminishes the structural benefits of our optical-flow-based densification.
For dense flow estimation, we employ LiteFlowNet3~\cite{hui2020liteflownet3}, an ultra-lightweight deep neural network, at a fixed step size of 7 keyframes.
Finally, during each training iteration of the Gaussian model, we sample from the most recent $k=25$ keyframes with a high uniform probability of 0.7, drawing from older keyframes with a 0.3 probability to mitigate forgetting.

\begin{figure*}
    \setlength{\tabcolsep}{3pt}

    \begin{center}
    \begin{tabular}{ @{} c c c c @{} }
     \includegraphics[width=0.24\textwidth]{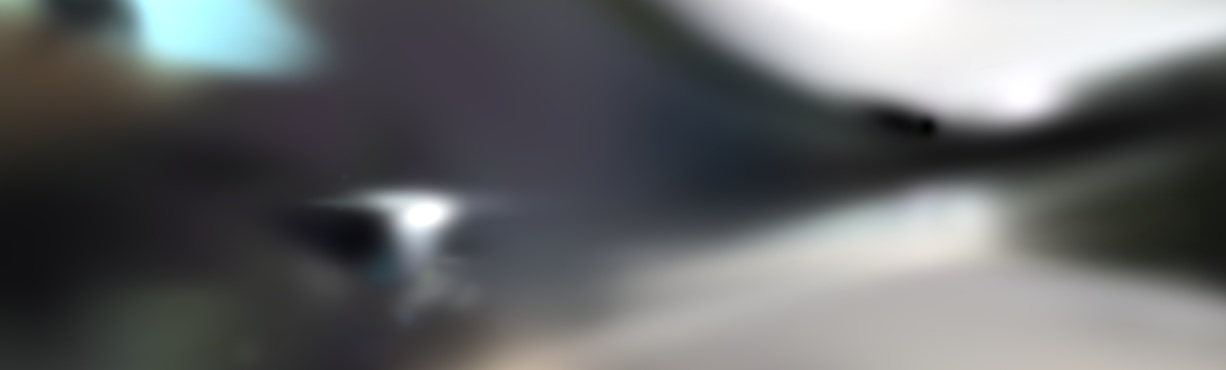} & \includegraphics[width=0.24\textwidth]{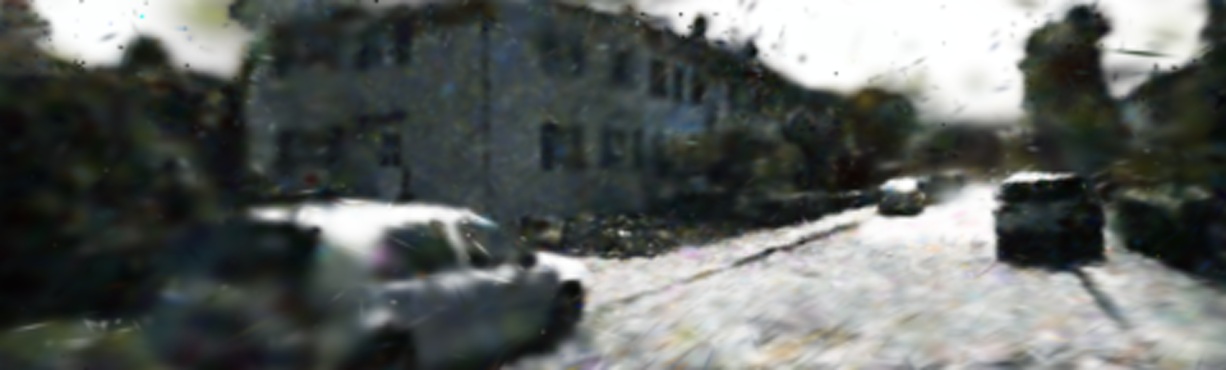} & \includegraphics[width=0.24\textwidth]{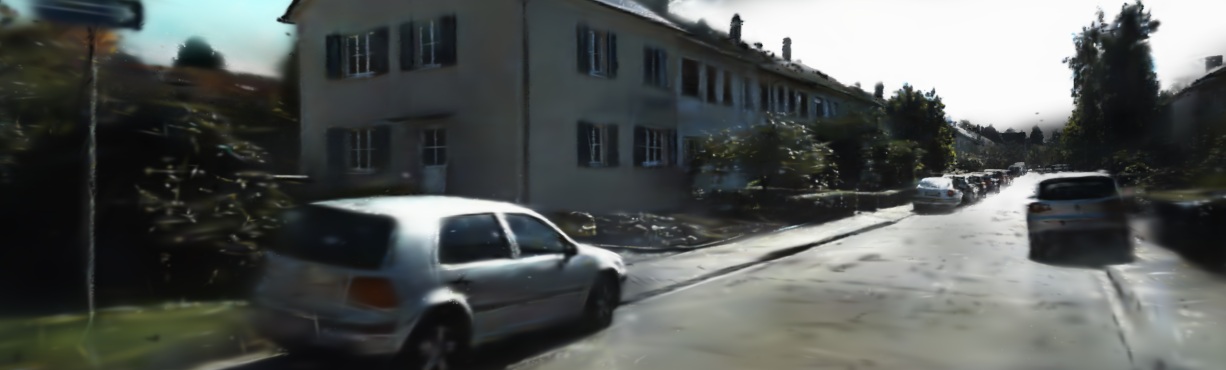} & \includegraphics[width=0.24\textwidth]{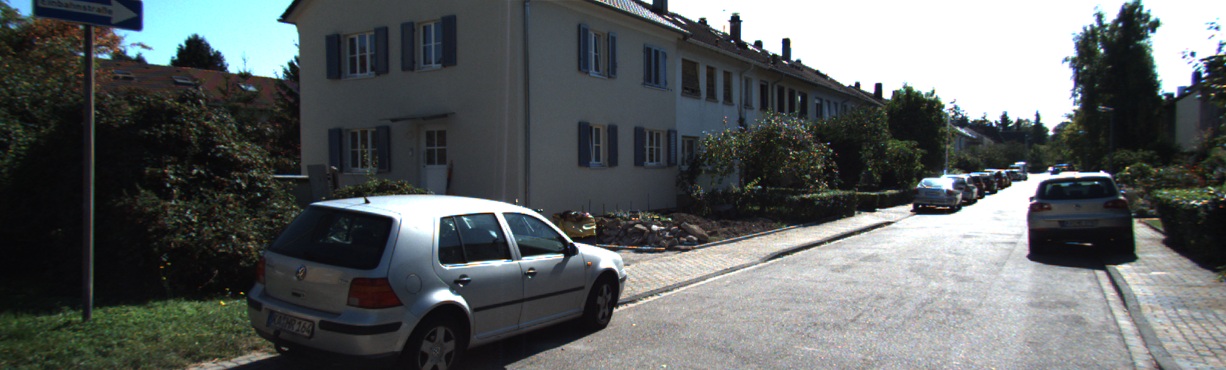} \\
     
     \includegraphics[width=0.24\textwidth]{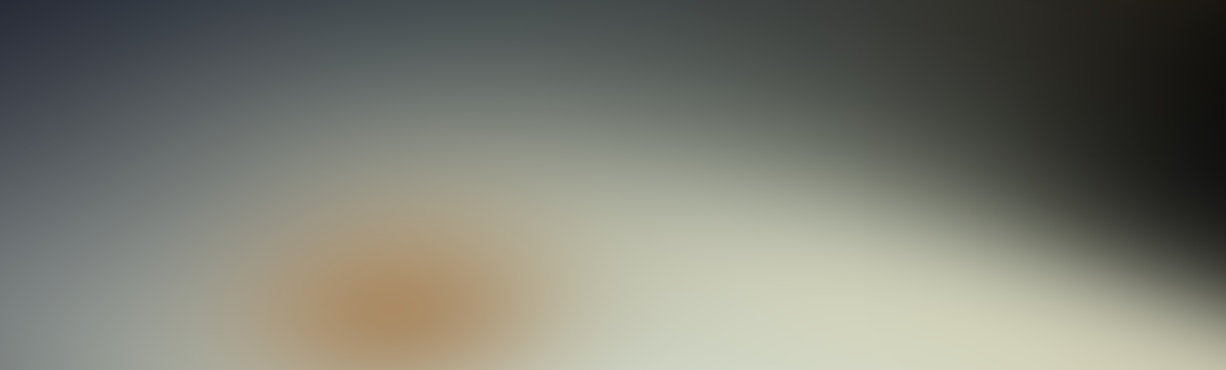} & \includegraphics[width=0.24\textwidth]{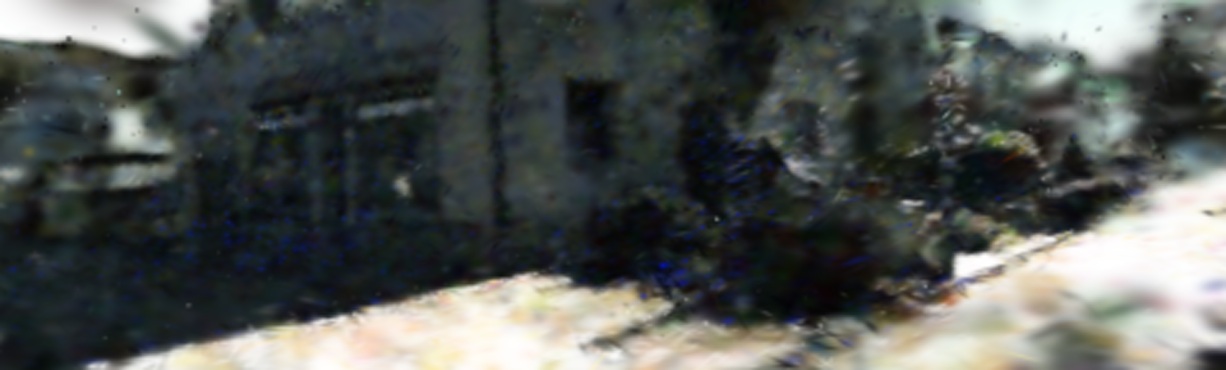} & \includegraphics[width=0.24\textwidth]{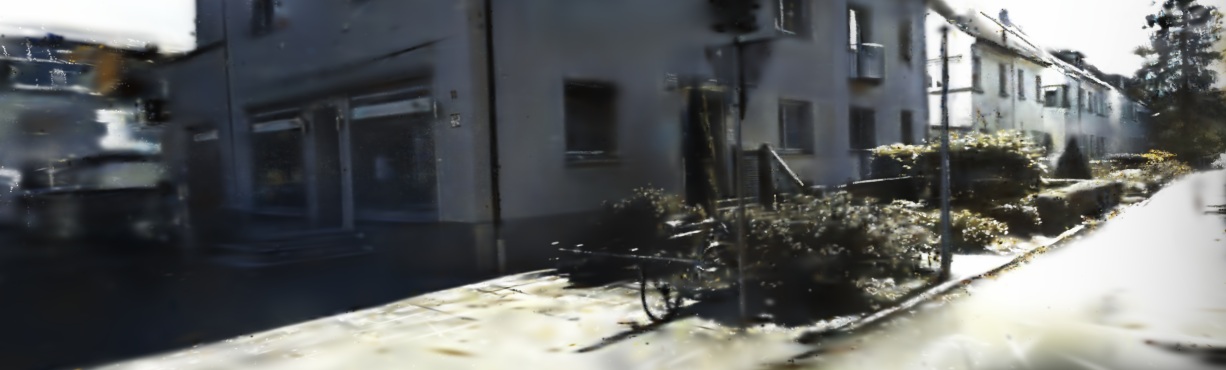} & \includegraphics[width=0.24\textwidth]{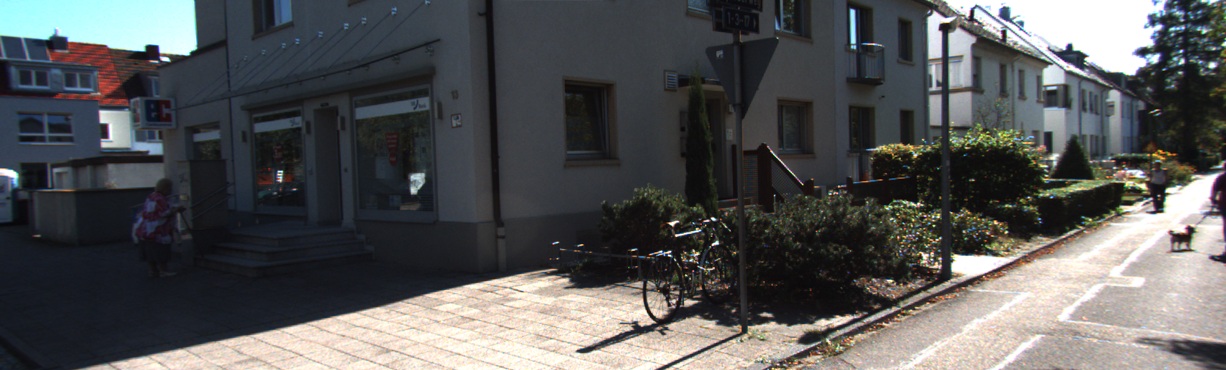} \\
     
     Out-of-Memory \vspace{0.25cm} & Out-of-Memory & \includegraphics[width=0.24\textwidth]{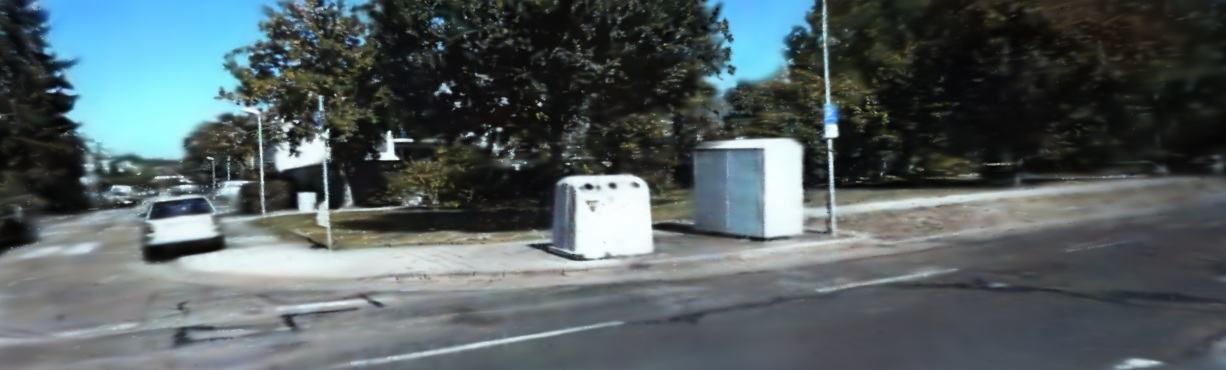} & \includegraphics[width=0.24\textwidth]{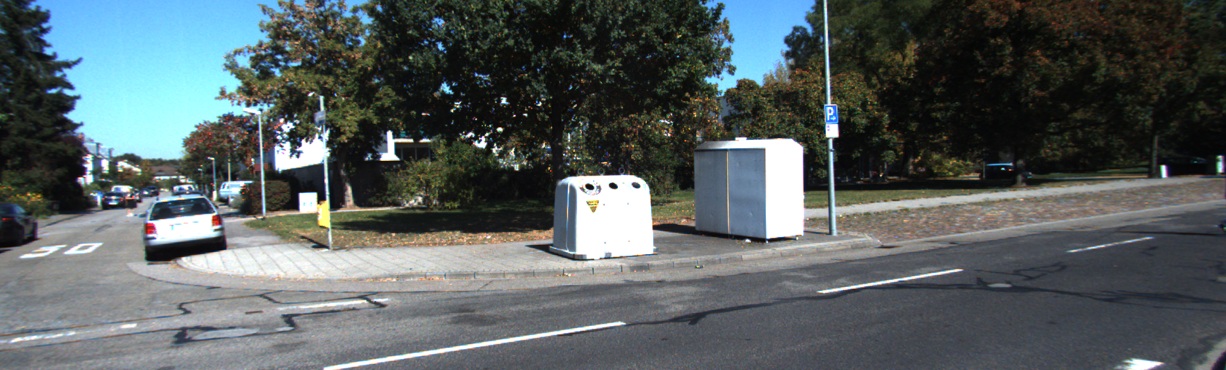} \\
     (a) PhotoSLAM & (b) GigaSLAM & (c) Ours & (d) Ground Truth
    \end{tabular}
    \end{center}
    
    \vspace{-0.6\baselineskip}
    \caption{\textbf{Qualitative rendering comparison on KITTI Odometry Seq.~\texttt{08}:} Rows correspond to frames 500, 2000, and 4000. PhotoSLAM (a) exhibits significant rendering degradation over time. PhotoSLAM (a) and GigaSLAM (b) terminate due to memory exhaustion after 2000 and 3200 frames, respectively. MonoGS fails on this sequence and is not included. In contrast, our method maintains consistent rendering quality throughout all 4071 frames.}
    \label{fig:results_renderingcomparison}
\end{figure*}

\subsection{Photometric Evaluation}

We evaluate our system's photometric performance across several dataset sequences: 11 from KITTI Odometry~\cite{geiger2013:kitti}, and 3 each from Oxford RobotCar~\cite{maddern2017:RobotCarDataset} and M\'alaga~\cite{blancoclaraco2014:malaga}, with quantitative results reported in Tables~\ref{table:results_qualitymetricsmain} and~\ref{table:results_qualitymetricsextra} and rendering comparison in Fig.~\ref{fig:results_renderingcomparison}. Overall, our method clearly surpasses both PhotoSLAM and GigaSLAM (pre-opt), achieving noticeably higher reconstruction quality across all metrics. Compared to GigaSLAM, the next best performer, our approach yields average improvements in (PSNR, SSIM, LPIPS) of (11.6\%, 9.2\%, 28.0\%) on KITTI, (26.6\%, 15.7\%, 35.3\%) on Oxford RobotCar, and (15.7\%, 6.3\%, 10.5\%) on M\'alaga. While GigaSLAM's post-optimization boosts photometric scores, this strictly offline refinement sacrifices real-time capability and can unpredictably fail, notably degrading sequence \texttt{00} with needle-like Gaussians that sharply reduce all quality metrics.

\begin{table}[t]
\setlength{\tabcolsep}{3pt}

\settowidth{\wexp}{Static Line Trackletsssssssssssss}
\newcolumntype{C}{>{\centering\arraybackslash}p{\dimexpr(\wexp-2\tabcolsep)/2}}

\settowidth{\xexp}{15.020}
\newcolumntype{D}{>{\centering\arraybackslash}p{\dimexpr(\xexp)}}

\settowidth{\yexp}{0.41000}
\newcolumntype{E}{>{\raggedleft\arraybackslash}p{\dimexpr(\yexp)}}

\centering
\caption{Performance metrics (FPS, peak GPU memory allocation, number of scene primitives) on the KITTI Odometry Dataset. Scene primitives are Gaussians for PhotoSLAM and anchors for the other methods. $^\times$: Due to Out-of-Memory limitations, PhotoSLAM has its reported results for 2000 frames and GigaSLAM for 3200 frames.}
\vspace{-0.75\baselineskip}
\resizebox{0.49\textwidth}{!}{%
\begin{tabular}{@{} l @{}p{0.15cm} C || *{11}{E} || E }
\toprule[1.5pt] \addlinespace[\topTableSpacing]
\textbf{Method} & & \textbf{Sequences} $\rightarrow$ & \multicolumn{1}{c}{\texttt{00}} & \multicolumn{1}{c}{\texttt{01}} & \multicolumn{1}{c}{\texttt{02}} &  \multicolumn{1}{c}{\texttt{03}} & \multicolumn{1}{c}{\texttt{04}} & \multicolumn{1}{c}{\texttt{05}} & \multicolumn{1}{c}{\texttt{06}} & \multicolumn{1}{c}{\texttt{07}} & \multicolumn{1}{c}{\texttt{08}} & \multicolumn{1}{c}{\texttt{09}} & \multicolumn{1}{c||}{\texttt{10}} & \multicolumn{1}{c}{\textbf{Avg.}} \\
  
\textit{Total frames} & & & \multicolumn{1}{c}{\textit{4541}} & \multicolumn{1}{c}{\textit{1101}} & \multicolumn{1}{c}{\textit{4661}} & \multicolumn{1}{c}{\textit{801}} & \multicolumn{1}{c}{\textit{271}} & \multicolumn{1}{c}{\textit{2761}} & \multicolumn{1}{c}{\textit{1101}} & \multicolumn{1}{c}{\textit{1101}} & \multicolumn{1}{c}{\textit{4071}} & \multicolumn{1}{c}{\textit{1591}} & \multicolumn{1}{c||}{\textit{1201}} \\ \addlinespace[\topTableSpacing] \hline \addlinespace[\topTableSpacing]

 & & FPS                 & \textcolor{softgray}{\textit{$\sim$10}$^{\times}$}                  & $\sim$10\phantom{$^{\times}$}       & \textcolor{softgray}{\textit{$\sim$10}$^{\times}$}     & $\sim$10\phantom{$^{\times}$}          & $\sim$10\phantom{$^{\times}$}          & \textcolor{softgray}{\textit{$\sim$10}$^{\times}$}         & $\sim$10\phantom{$^{\times}$}          & $\sim$10\phantom{$^{\times}$} &  \textcolor{softgray}{\textit{$\sim$10}$^{\times}$}         & $\sim$10\phantom{$^{\times}$} & $\sim$10\phantom{$^{\times}$}          & \textbf{$\sim$10}\phantom{$^{\times}$}    \\

PhotoSLAM & & GPU (GiB)                 & \textcolor{softgray}{\textit{25.0}$^{\times}$}                  & 8.2\phantom{$^{\times}$}       & \textcolor{softgray}{\textit{29.1}$^\times$}     & 6.3\phantom{$^{\times}$}           & 2.8\phantom{$^{\times}$}          &  \textcolor{softgray}{\textit{23.2}$^\times$}        & 9.7\phantom{$^{\times}$}         & 9.8\phantom{$^{\times}$}  & \textcolor{softgray}{\textit{30.9}$^\times$} & 17.9\phantom{$^{\times}$} &            11.0\phantom{$^{\times}$} & 15.8\phantom{$^{\times}$}   \\

 & & Primitives                 & \textcolor{softgray}{\textit{907k}$^{\times}$}                  & 347k\phantom{$^{\times}$}       & \textcolor{softgray}{\textit{820k}$^\times$}     & 371k\phantom{$^{\times}$}          & 101k\phantom{$^{\times}$}          & \textcolor{softgray}{\textit{718k}$^\times$}         & 308k\phantom{$^{\times}$}          & 496k\phantom{$^{\times}$}  & \textcolor{softgray}{\textit{877k}$^\times$}          & 620k\phantom{$^{\times}$} &           455k\phantom{$^{\times}$} & \textbf{547k}\phantom{$^{\times}$}  \\ \addlinespace[\topTableSpacing] \hline \addlinespace[\topTableSpacing]

 & & FPS                 &  \textcolor{softgray}{\textit{2.89}$^\times$}  &  4.84\phantom{$^{\times}$}      & \textcolor{softgray}{\textit{2.65}$^\times$}     &  3.76\phantom{$^{\times}$}         &  3.39\phantom{$^{\times}$}        & 2.69\phantom{$^{\times}$} & 2.76\phantom{$^{\times}$}         & 3.21\phantom{$^{\times}$}           & \textcolor{softgray}{\textit{2.94}$^\times$}          & 2.73\phantom{$^{\times}$} & 3.02\phantom{$^{\times}$}          & 3.17\phantom{$^{\times}$}    \\

GigaSLAM & & GPU (GiB)                 &  \textcolor{softgray}{\textit{31.1}$^\times$} & 12.2\phantom{$^{\times}$}       & \textcolor{softgray}{\textit{29.0}$^\times$}     & 12.8\phantom{$^{\times}$}          & 9.9\phantom{$^{\times}$}        & 23.8\phantom{$^{\times}$} & 14.4\phantom{$^{\times}$} & 14.4\phantom{$^{\times}$}  &     \textcolor{softgray}{\textit{26.4}$^\times$}      & 16.8\phantom{$^{\times}$} & 14.9\phantom{$^{\times}$}          &   18.7\phantom{$^{\times}$}  \\

(pre-opt) & & Primitives                 & \textcolor{softgray}{\textit{2811k}$^\times$} & 550k\phantom{$^{\times}$}       & \textcolor{softgray}{\textit{3075k}$^\times$}     & 548k\phantom{$^{\times}$}          & 223k\phantom{$^{\times}$}         & 2249k\phantom{$^{\times}$} & 925k\phantom{$^{\times}$} & 849k\phantom{$^{\times}$} & \textcolor{softgray}{\textit{2531k}$^\times$}          & 1638k\phantom{$^{\times}$} & 991k\phantom{$^{\times}$}          & 1490k\phantom{$^{\times}$}    \\ \addlinespace[\topTableSpacing] \hline \addlinespace[\topTableSpacing]

& & FPS    &  & 1.81\phantom{$^{\times}$}  &      &  &  1.45\phantom{$^{\times}$} &   & 4.49\phantom{$^{\times}$}  &  &   &  &    & \multicolumn{1}{c}{-}    \\

MonoGS & & GPU (GiB)  & \textcolor{softgray}{\textit{failed}}\phantom{$^{\times}$}  &  7.3\phantom{$^{\times}$}  & \textcolor{softgray}{\textit{failed}}\phantom{$^{\times}$}     &     \textcolor{softgray}{\textit{failed}}\phantom{$^{\times}$}       &   4.4\phantom{$^{\times}$} &  \textcolor{softgray}{\textit{failed}}\phantom{$^{\times}$}   &  9.9\phantom{$^{\times}$}  & \textcolor{softgray}{\textit{failed}}\phantom{$^{\times}$}  & \textcolor{softgray}{\textit{failed}}\phantom{$^{\times}$} & \textcolor{softgray}{\textit{failed}}\phantom{$^{\times}$} &    \textcolor{softgray}{\textit{failed}}\phantom{$^{\times}$}  &  \multicolumn{1}{c}{-} \\
 
& & Primitives                 &     &   36k\phantom{$^{\times}$} &   &   & 23k\phantom{$^{\times}$}          &   & 22k\phantom{$^{\times}$}          &  &    &  &   & \multicolumn{1}{c}{-} \\ \addlinespace[\topTableSpacing] \hline \addlinespace[\topTableSpacing]

 & & FPS & $\sim$10\phantom{$^{\times}$} & $\sim$10\phantom{$^{\times}$}      & $\sim$10\phantom{$^{\times}$} & $\sim$10\phantom{$^{\times}$} & $\sim$10\phantom{$^{\times}$}  & $\sim$10\phantom{$^{\times}$} & $\sim$10\phantom{$^{\times}$} & $\sim$10\phantom{$^{\times}$} & $\sim$10\phantom{$^{\times}$}  & $\sim$10\phantom{$^{\times}$} & $\sim$10\phantom{$^{\times}$} & \textbf{$\sim$10}\phantom{$^{\times}$}     \\
 
Ours & & GPU (GiB) & 17.6\phantom{$^{\times}$} & 7.2\phantom{$^{\times}$} & 15.6\phantom{$^{\times}$} & 7.8\phantom{$^{\times}$} & 5.6\phantom{$^{\times}$} & 15.2\phantom{$^{\times}$} & 8.1\phantom{$^{\times}$} & 8.8\phantom{$^{\times}$} & 21.8\phantom{$^{\times}$} & 11.4\phantom{$^{\times}$} & 8.1\phantom{$^{\times}$} & \textbf{11.6}\phantom{$^{\times}$}   \\

 & & Primitives & 1388k\phantom{$^{\times}$} & 112k\phantom{$^{\times}$} & 995k\phantom{$^{\times}$} & 547k\phantom{$^{\times}$} & 130k\phantom{$^{\times}$} & 1029k\phantom{$^{\times}$} & 422k\phantom{$^{\times}$} & 787k\phantom{$^{\times}$} & 1681k\phantom{$^{\times}$} & 751k\phantom{$^{\times}$} & 443k\phantom{$^{\times}$} & 753k\phantom{$^{\times}$}   \\ \addlinespace[\topTableSpacing] \hline \hline \addlinespace[\topTableSpacing]

 & & FPS                 &  0.32\phantom{$^{\times}$}                 & 1.08\phantom{$^{\times}$}       & 0.28\phantom{$^{\times}$}     &  0.79\phantom{$^{\times}$}         & 0.99\phantom{$^{\times}$}         &     0.36\phantom{$^{\times}$}     & 0.54\phantom{$^{\times}$}          & 0.58\phantom{$^{\times}$} & 0.36\phantom{$^{\times}$}          & 0.39\phantom{$^{\times}$} & 0.54\phantom{$^{\times}$}          &  0.57\phantom{$^{\times}$}   \\
 
GigaSLAM & & GPU (GiB) & 31.2\phantom{$^{\times}$} & 12.3\phantom{$^{\times}$} & 29.1\phantom{$^{\times}$} & 12.8\phantom{$^{\times}$} & 10.0\phantom{$^{\times}$} & 23.9\phantom{$^{\times}$} & 14.6\phantom{$^{\times}$} & 14.5\phantom{$^{\times}$} & 26.6\phantom{$^{\times}$} & 16.8\phantom{$^{\times}$} & 15.0\phantom{$^{\times}$} & 18.8\phantom{$^{\times}$}   \\

(post-opt)  & & Primitives                 & 2811k\phantom{$^{\times}$} & 550k\phantom{$^{\times}$}       & 3076k\phantom{$^{\times}$}     & 561k\phantom{$^{\times}$}          & 223k\phantom{$^{\times}$}           &     2247k\phantom{$^{\times}$}     & 925k\phantom{$^{\times}$}          & 849k\phantom{$^{\times}$} & 2531k\phantom{$^{\times}$}           & 1638k\phantom{$^{\times}$} & 991k\phantom{$^{\times}$}          & 1491k\phantom{$^{\times}$}    \\ \addlinespace[\bottomTableSpacing] \bottomrule[1.5pt]

\end{tabular}
}
\vspace{-0.25\baselineskip}
\label{table:results_performancemetricsmain}

\end{table}

\subsection{Evaluation of Computational Performance}

Our system and PhotoSLAM consistently operate in real-time across all evaluated datasets (KITTI: 10 FPS, Oxford RobotCar: 16 FPS, M\'alaga: 20 FPS). In contrast, GigaSLAM (pre-opt) fails to operate in real time, bottlenecked by heavy neural networks for depth and feature matching, as well as costly loop closures (3x slower on KITTI; see Table~\ref{table:results_performancemetricsmain}).

In terms of memory, our system exhibits the lowest peak GPU memory usage, scaling efficiently despite the constant 3\,GiB allocated on the GPU for loading the optical flow model. This efficiency enables processing of long sequences without interruption, while the other methods encounter Out-of-Memory failures on our hardware (Seq. \texttt{00}, \texttt{02}, \texttt{05}, \texttt{08}), requiring sequence truncation for evaluation. GigaSLAM's high memory footprint stems from loading multiple networks and maintaining a large number of anchors. Our approach, despite using more primitives than PhotoSLAM, still shows lower GPU memory consumption, which is primarily due to the more compact representation of the structured anchors.

\begin{figure}
    \centering
    \includegraphics[width=0.238\textwidth]{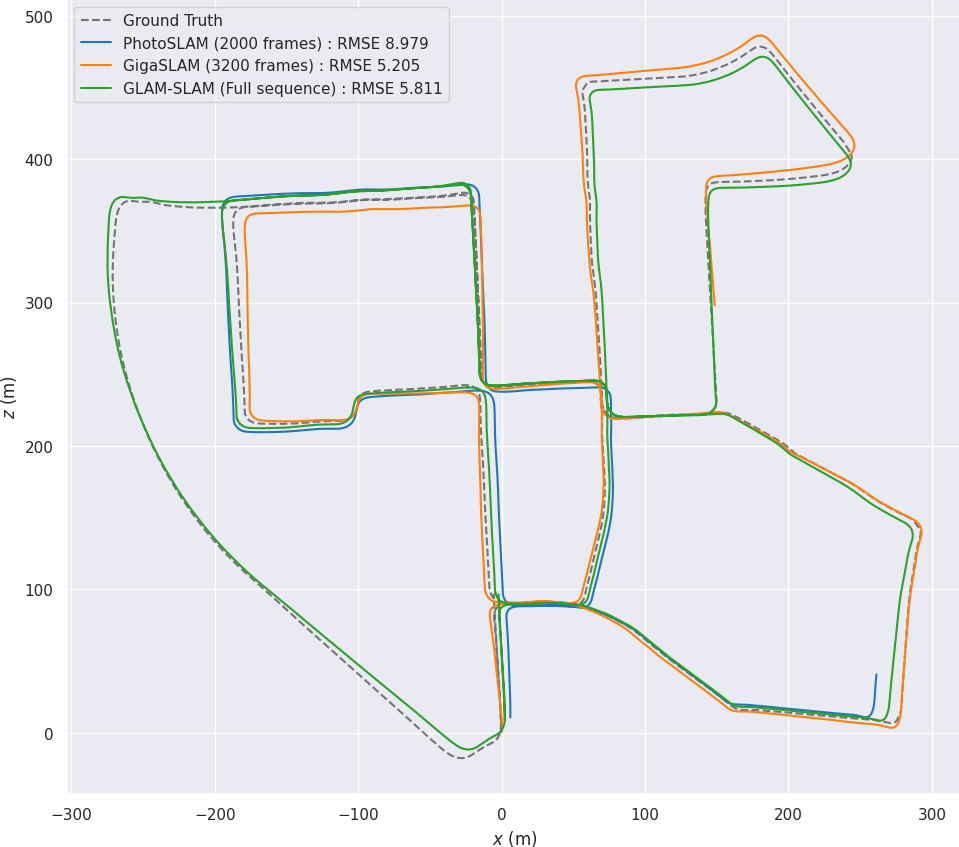} \hfill \includegraphics[width=0.238\textwidth]{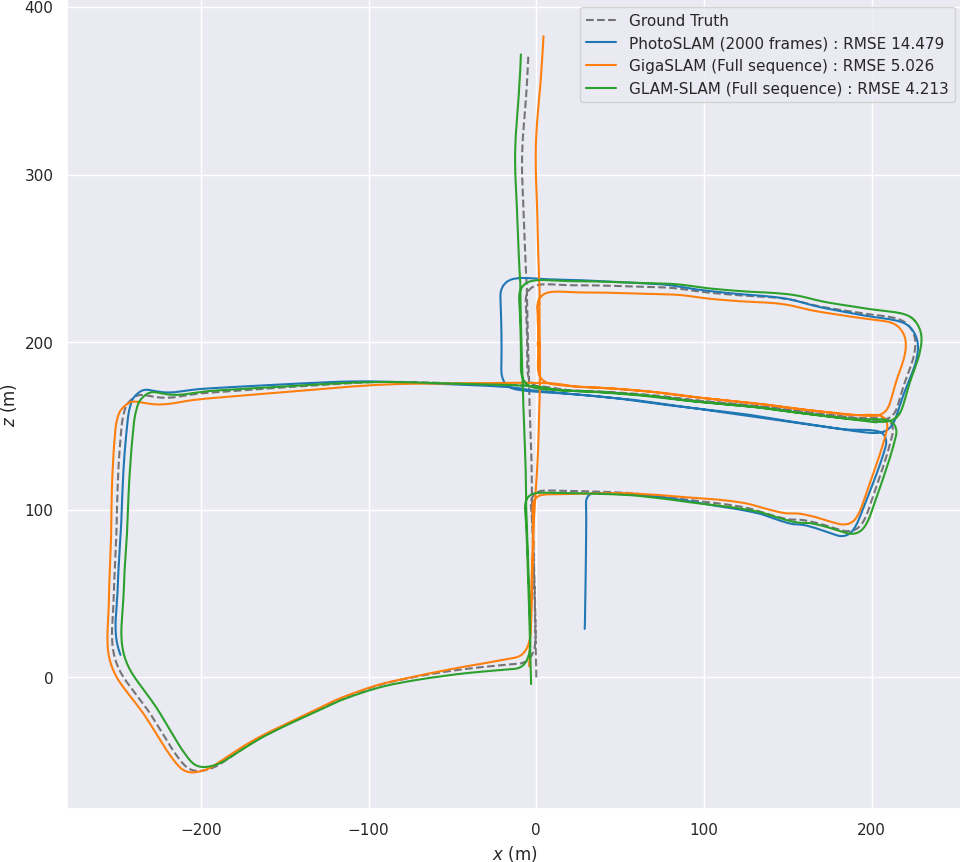}
    
    \vspace{-0.5\baselineskip}
    \caption{\textbf{Trajectory comparison on KITTI Seqs. \texttt{00}, \texttt{05}:} GLAM-SLAM is the only method that completes the entire 4541 and 2761-frame sequences.}
    \label{fig:ate_qualitative_trajectory_comparison}
\end{figure}

\begin{table}
\setlength{\tabcolsep}{5pt}

\settowidth{\wexp}{Static Line Trackletsssssssssssss}
\newcolumntype{C}{>{\centering\arraybackslash}p{\dimexpr(\wexp-2\tabcolsep)/2}}

\settowidth{\xexp}{15.020}
\newcolumntype{D}{>{\centering\arraybackslash}p{\dimexpr(\xexp)}}

\settowidth{\yexp}{0.4100}
\newcolumntype{E}{>{\raggedleft\arraybackslash}p{\dimexpr(\yexp)}}

\centering
\caption{Absolute Trajectory Error (ATE RMSE [m]) on the KITTI Odometry Dataset. $^\times$: Due to Out-of-Memory limitations, PhotoSLAM reports results for 2000 frames and GigaSLAM for 3200 frames.}
\vspace{-0.75\baselineskip}
\scriptsize\resizebox{0.485\textwidth}{!}{%
\begin{tabular}{@{} l @{}p{0.15cm} || *{11}{E} }
\toprule[1.5pt] \addlinespace[\topTableSpacing]
    \textbf{Method} & & \multicolumn{1}{c}{\texttt{00}} & \multicolumn{1}{c}{\texttt{01}} & \multicolumn{1}{c}{\texttt{02}} &  \multicolumn{1}{c}{\texttt{03}} & \multicolumn{1}{c}{\texttt{04}} & \multicolumn{1}{c}{\texttt{05}} & \multicolumn{1}{c}{\texttt{06}} & \multicolumn{1}{c}{\texttt{07}} & \multicolumn{1}{c}{\texttt{08}} & \multicolumn{1}{c}{\texttt{09}} & \multicolumn{1}{c}{\texttt{10}} \\
  
\textit{Total frames} &  & \multicolumn{1}{c}{\textit{4541}} & \multicolumn{1}{c}{\textit{1101}} & \multicolumn{1}{c}{\textit{4661}} & \multicolumn{1}{c}{\textit{801}} & \multicolumn{1}{c}{\textit{271}} & \multicolumn{1}{c}{\textit{2761}} & \multicolumn{1}{c}{\textit{1101}} & \multicolumn{1}{c}{\textit{1101}} & \multicolumn{1}{c}{\textit{4071}} & \multicolumn{1}{c}{\textit{1591}} & \multicolumn{1}{c}{\textit{1201}} \\ \addlinespace[\topTableSpacing] \hline \addlinespace[\topTableSpacing]

PhotoSLAM & &  \textcolor{softgray}{\textit{8.98$^\times$}} & 518.51\phantom{.} & \textcolor{softgray}{\textit{10.57$^\times$}} & \textbf{0.76}\phantom{ss} & 1.43\phantom{ss} & \textcolor{softgray}{\textit{14.48$^\times$}} & 16.48\phantom{s} & 2.50\phantom{si} & \textcolor{softgray}{\textit{44.78$^\times$}} & 6.90\phantom{si} & 6.68\phantom{si} \\ \addlinespace[\topTableSpacing] \hline \addlinespace[\topTableSpacing]

GigaSLAM & & \textcolor{softgray}{\textit{5.20$^\times$}} & \textbf{189.10}\phantom{.} & \textcolor{softgray}{\textit{6.75$^\times$}} & 1.29\phantom{ss} & 1.59\phantom{ss} & 5.03\phantom{$^{\times}$} & \textbf{1.37}\phantom{s} & 3.00\phantom{si} & \textcolor{softgray}{\textit{3.89$^\times$}} & \textbf{3.85}\phantom{si} & \textbf{2.28}\phantom{si}  \\ \addlinespace[\topTableSpacing] \hline \addlinespace[\topTableSpacing]

MonoGS & & \textcolor{softgray}{\textit{failed}}\phantom{$^{\times}$} & 402.64\phantom{.} & \textcolor{softgray}{\textit{failed}}\phantom{$^{\times}$} & \textcolor{softgray}{\textit{failed}}\phantom{ss} & 16.19\phantom{ss} & \textcolor{softgray}{\textit{failed}}\phantom{$^{\times}$} & 57.24\phantom{s} & \textcolor{softgray}{\textit{failed}}\phantom{si} & \textcolor{softgray}{\textit{failed}}\phantom{$^{\times}$} & \textcolor{softgray}{\textit{failed}}\phantom{si} & \textcolor{softgray}{\textit{failed}}\phantom{si} \\ \addlinespace[\topTableSpacing] \hline \addlinespace[\topTableSpacing]

Ours & & 5.81\phantom{$^{\times}$} & 364.20\phantom{.} & 17.61\phantom{$^{\times}$} & 1.00\phantom{ss} & \textbf{1.27}\phantom{ss} & \textbf{4.21}\phantom{$^{\times}$} & 15.76\phantom{s} & \textbf{2.19}\phantom{si} & 46.07\phantom{$^{\times}$} & 8.39\phantom{si} & 5.64\phantom{si} \\ \addlinespace[\bottomTableSpacing] \bottomrule[1.5pt]

\end{tabular}
}
\vspace{-1.2\baselineskip}
\label{table:results_ate}

\end{table}

\begin{figure*}[ht]
    \centering
    \begin{minipage}{\textwidth}
        \centering
        \includegraphics[width=\textwidth]{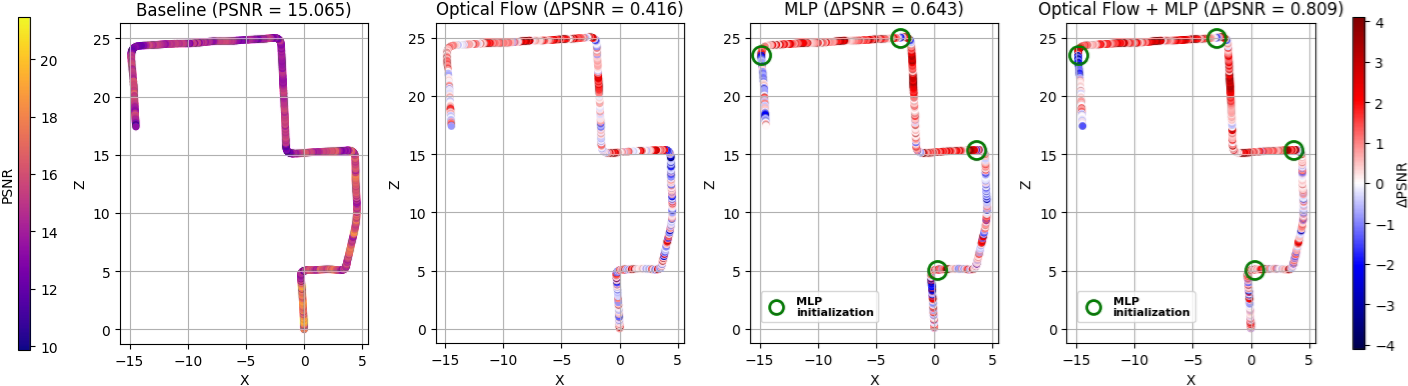}
    \end{minipage}

    \caption{\textbf{Per-keyframe quality analysis along the X--Z trajectory:} We visualize absolute PSNR (first plot) and relative improvement $\Delta\text{PSNR}$ (remaining plots) for the ablation on KITTI Seq.~\texttt{00} (first 1000 frames). The average PSNR and $\Delta\text{PSNR}$ values reported match those presented in Table~\ref{table:results_qualityandperformancemetricsablations}.}
    \label{fig:supplementary_kitti00_psnr_deltapsnr_xyz}
\end{figure*}

\subsection{ATE Evaluation}

Quantitative results (Table~\ref{table:results_ate}) show our system achieves real-time tracking comparable to PhotoSLAM due to their shared ORB-SLAM-based frontend.
However, both PhotoSLAM and GigaSLAM face severe scalability limits, exhausting GPU memory on sequences exceeding 2000 and 3200 frames, respectively.
MonoGS crashes after a few hundred frames on most sequences.
While GigaSLAM achieves lower average ATE (3.4m vs 10.8m excluding the mutual catastrophic failure in Seq. \texttt{01}), this comes at the cost of non-real-time performance ($\sim$3 FPS) and a heavy backend that depletes 32GB VRAM.
In contrast, our lightweight system maintains acceptable tracking performance and scales robustly to unbounded, long-sequence outdoor trajectories without resource depletion. Fig.~\ref{fig:ate_qualitative_trajectory_comparison} compares trajectories on Seqs.~\texttt{00} and~\texttt{05}, where our approach maintains stable estimates while PhotoSLAM and GigaSLAM terminate early due to Out-of-Memory errors.

\subsection{Ablation Studies}

We evaluate our main contributions through a detailed ablation study of up to 1000 frames of each KITTI sequence (except Seq.~\texttt{00} because ORB-SLAM2 loses tracking), as presented in Table~\ref{table:results_qualityandperformancemetricsablations}.

\textbf{Flow Densification:} Incorporating optical-flow--based correspondences yields a substantial improvement in photometric reconstruction metrics across all KITTI sequences, along with qualitative enhancements in the rendered images, particularly in sparsely tracked regions, as seen in Fig.~\ref{fig:supplementary_kitti00_qualitative_295}. This results in a 32\% richer representation of Gaussian primitives, which brings the average memory footprint to 6.5\,GiB (up from 1.8\,GiB), with the optical flow model inherently accounting for a flat $\sim$3\,GiB of this total. As shown in Table~\ref{table:results_qualityandperformancemetricsablations}, shifting computational resources to optical flow results in a modest decrease in raw Gaussian optimization iterations (10.6k vs. 11.7k). This is a highly advantageous trade-off: integrating geometric priors from the flow estimation accelerates convergence, yielding higher-quality results with fewer iterations while strictly maintaining real-time operation.

\textbf{Localized MLPs:} Likewise, introducing localized MLPs in distinct regions yields a noticeable improvement in photometric reconstruction metrics (see Fig.~\ref{fig:supplementary_kitti00_psnr_deltapsnr_xyz} for $\Delta$PSNR improvement) and results in a substantial increase in the number of Gaussian primitives (approximately 30\%). Notably, in six sequences, the isolated use of multiple MLPs yields better performance than the optical flow module, achieving this while maintaining the exact VRAM footprint of the baseline.
We note that although explicit cross-partition consistency is not enforced, the resulting photometric fluctuations at spatial boundaries do not perceptually degrade the rendering quality.

It should be emphasized that these improvements are \textit{complementary}: enabling both leads to further gains, as demonstrated in Table~\ref{table:results_qualityandperformancemetricsablations}.
We conduct all experiments of Tables~\ref{table:results_qualitymetricsmain} and~\ref{table:results_performancemetricsmain} with both enhancements enabled.

\begin{table}
\setlength{\tabcolsep}{5pt}

\settowidth{\wexp}{Static Line Trackletssssssssssssss}
\newcolumntype{C}{>{\raggedleft\arraybackslash}p{\dimexpr(\wexp-2\tabcolsep)/2}}

\settowidth{\xexp}{15.020}
\newcolumntype{D}{>{\centering\arraybackslash}p{\dimexpr(\xexp)}}

\settowidth{\yexp}{0.4100}
\newcolumntype{E}{>{\raggedleft\arraybackslash}p{\dimexpr(\yexp)}}

\centering
\caption{Ablation study of photometric reconstruction quality (PSNR, SSIM, LPIPS) and performance metrics (FPS, peak GPU memory allocation, number of scene primitives) on the KITTI Odometry dataset. $^\text{F}$: ORB-SLAM2 tracking has failed.}
\vspace{-0.75\baselineskip}
\resizebox{0.485\textwidth}{!}{%
\begin{tabular}{@{} l @{}p{0.15cm} C || *{11}{E} || E }
\toprule[1.5pt] \addlinespace[\topTableSpacing]
\textbf{Method} & & \multicolumn{1}{c||}{\textbf{Sequences} $\rightarrow$} & \multicolumn{1}{c}{\texttt{00}} & \multicolumn{1}{c}{\texttt{01}$^\text{F}$} & \multicolumn{1}{c}{\texttt{02}} & \multicolumn{1}{c}{\texttt{03}} & \multicolumn{1}{c}{\texttt{04}} & \multicolumn{1}{c}{\texttt{05}} & \multicolumn{1}{c}{\texttt{06}} & \multicolumn{1}{c}{\texttt{07}} & \multicolumn{1}{c}{\texttt{08}} & \multicolumn{1}{c}{\texttt{09}} & \multicolumn{1}{c||}{\texttt{10}} & \multicolumn{1}{c}{\textbf{Avg.}} \\
  
\textit{Total frames} &  &  & \multicolumn{1}{c}{\textit{1000}} & \multicolumn{1}{c}{\textit{-}} & \multicolumn{1}{c}{\textit{1000}} & \multicolumn{1}{c}{\textit{801}} & \multicolumn{1}{c}{\textit{271}} & \multicolumn{1}{c}{\textit{1000}} & \multicolumn{1}{c}{\textit{1000}} & \multicolumn{1}{c}{\textit{1000}} & \multicolumn{1}{c}{\textit{1000}} & \multicolumn{1}{c}{\textit{1000}} & \multicolumn{1}{c||}{\textit{1000}} \\ \addlinespace[\topTableSpacing] \hline \addlinespace[\topTableSpacing]

  & & PSNR $\uparrow$\phantom{ssi.}                 &   15.065                &    \multicolumn{1}{c}{-}      & 15.039                  &  19.044         & 20.851          & 19.020          & 18.618         & 18.852  & 15.953  & 18.882 & 16.377  & 17.770            \\

Baseline & & SSIM $\uparrow$\phantom{ssi.}                &  \phantom{6}0.816                &    \multicolumn{1}{c}{-}       & \phantom{6}0.785                  & \phantom{6}0.877          & \phantom{6}0.907          & \phantom{6}0.882 & \phantom{6}0.867         & \phantom{6}0.881  & \phantom{6}0.798 & \phantom{6}0.876         & \phantom{6}0.825  & \phantom{6}0.851 \\

 & & LPIPS $\downarrow$\phantom{ssi.}                & \phantom{6}0.671                  &    \multicolumn{1}{c}{-}       & \phantom{6}0.751                & \phantom{6}0.483          & \phantom{6}0.451          & \phantom{6}0.443         & \phantom{6}0.539          & \phantom{6}0.434 & \phantom{6}0.640  & \phantom{6}0.539          & \phantom{6}0.642 & \phantom{6}0.559    \\ \addlinespace[\topTableSpacing] \hline \addlinespace[\topTableSpacing]

  & & PSNR $\uparrow$\phantom{ssi.}                 & 15.481    &     \multicolumn{1}{c}{-}      & 15.215                  & 20.844          & \textbf{21.390}         & 19.206 & 19.238  & 19.517  & 16.452 & 19.341 & 16.513 & 18.320              \\

Optical Flow & & SSIM $\uparrow$\phantom{ssi.}                & \phantom{6}0.826   &      \multicolumn{1}{c}{-}     & \phantom{6}0.794                  & \phantom{6}0.912          & \phantom{6}\textbf{0.920}         & \phantom{6}0.884 & \phantom{6}0.879 & \phantom{6}0.894            & \phantom{6}0.811 & \phantom{6}0.884         & \phantom{6}0.830 & \phantom{6}0.863 \\

& & LPIPS $\downarrow$\phantom{ssi.}                & \phantom{6}0.646  &     \multicolumn{1}{c}{-}      & \phantom{6}0.728                 & \phantom{6}0.376         & \phantom{6}\textbf{0.411}         & \phantom{6}0.433 & \phantom{6}0.496         & \phantom{6}0.391         & \phantom{6}0.608 & \phantom{6}0.512           & \phantom{6}0.638 & \phantom{6}0.524    \\ \addlinespace[\topTableSpacing] \hline \addlinespace[\topTableSpacing]

  & & PSNR $\uparrow$\phantom{ssi.} & 15.709 & \multicolumn{1}{c}{-} & 15.593 & 19.966 & 20.851 & 19.185 & 20.261 & 19.525  & 16.715 & 19.421 & 16.650 & 18.388              \\

MLP & & SSIM $\uparrow$\phantom{ssi.} & \phantom{6}0.837 & \multicolumn{1}{c}{-} & \phantom{6}0.811 & \phantom{6}0.896 & \phantom{6}0.907  & \phantom{6}0.887 & \phantom{6}0.905  & \phantom{6}0.896  & \phantom{6}0.823 & \phantom{6}0.886 & \phantom{6}0.833 & \phantom{6}0.868 \\

 & & LPIPS $\downarrow$\phantom{ssi.} & \phantom{6}0.641  &     \multicolumn{1}{c}{-}      & \phantom{6}0.720                & \phantom{6}0.437          & \phantom{6}0.451          & \phantom{6}0.441         & \phantom{6}0.453          & \phantom{6}0.412 & \phantom{6}0.599 & \phantom{6}0.524          & \phantom{6}0.636 & \phantom{6}0.531    \\ \addlinespace[\topTableSpacing] \hline \addlinespace[\topTableSpacing]

  & & PSNR $\uparrow$\phantom{ssi.}                 & \textbf{15.874}                  &      \multicolumn{1}{c}{-}     & \textbf{15.806}  & \textbf{20.870}          & \textbf{21.390}          & \textbf{19.613}         & \textbf{20.756}         & \textbf{19.925}  & \textbf{16.881}  & \textbf{19.772}  & \textbf{17.115} & \textbf{18.800}              \\

Optical Flow + & & SSIM $\uparrow$\phantom{ssi.}                & \phantom{6}\textbf{0.838}                 &   \multicolumn{1}{c}{-}        & \phantom{6}\textbf{0.817} & \phantom{6}\textbf{0.913}          & \phantom{6}\textbf{0.920}          & \phantom{6}\textbf{0.895} & \phantom{6}\textbf{0.913}         & \phantom{6}\textbf{0.904}  & \phantom{6}\textbf{0.829} & \phantom{6}\textbf{0.890}         & \phantom{6}\textbf{0.845} & \phantom{6}\textbf{0.876} \\

MLP & & LPIPS $\downarrow$\phantom{ssi.}                & \phantom{6}\textbf{0.632}                  & \multicolumn{1}{c}{-}  & \phantom{6}\textbf{0.701} & \phantom{6}\textbf{0.373}           & \phantom{6}\textbf{0.411}          & \phantom{6}\textbf{0.416}         & \phantom{6}\textbf{0.416}          & \phantom{6}\textbf{0.371} & \phantom{6}\textbf{0.587} & \phantom{6}\textbf{0.497}          & \phantom{6}\textbf{0.608} & \phantom{6}\textbf{0.501}    \\ \addlinespace[\bottomTableSpacing] \bottomrule[1.5pt]

\end{tabular}
}

\vspace{\baselineskip}

\resizebox{0.485\textwidth}{!}{%
\begin{tabular}{@{} l @{}p{0.15cm} C || *{11}{E} || E }
\toprule[1.5pt] \addlinespace[\topTableSpacing]
\textbf{Method} & & \textbf{Sequences} $\rightarrow$ & \multicolumn{1}{c}{\texttt{00}} & \multicolumn{1}{c}{\texttt{01}$^\text{F}$} & \multicolumn{1}{c}{\texttt{02}} & \multicolumn{1}{c}{\texttt{03}} & \multicolumn{1}{c}{\texttt{04}} & \multicolumn{1}{c}{\texttt{05}} & \multicolumn{1}{c}{\texttt{06}} & \multicolumn{1}{c}{\texttt{07}} & \multicolumn{1}{c}{\texttt{08}} & \multicolumn{1}{c}{\texttt{09}} & \multicolumn{1}{c||}{\texttt{10}} & \multicolumn{1}{c}{\textbf{Avg.}} \\

\textit{Total frames} &  &  & \multicolumn{1}{c}{\textit{1000}} & \multicolumn{1}{c}{\textit{-}} & \multicolumn{1}{c}{\textit{1000}} & \multicolumn{1}{c}{\textit{801}} & \multicolumn{1}{c}{\textit{271}} & \multicolumn{1}{c}{\textit{1000}} & \multicolumn{1}{c}{\textit{1000}} & \multicolumn{1}{c}{\textit{1000}} & \multicolumn{1}{c}{\textit{1000}} & \multicolumn{1}{c}{\textit{1000}} & \multicolumn{1}{c||}{\textit{1000}} \\ \addlinespace[\topTableSpacing] \hline \addlinespace[\topTableSpacing]

 & & \multicolumn{1}{c||}{GPU (GiB)} & 1.9\phantom{S} & \multicolumn{1}{c}{-} & 1.3\phantom{S} & 1.7\phantom{S} & 1.4\phantom{S} & 1.8\phantom{S} & 1.8\phantom{S} & 2.3\phantom{S} & 1.7\phantom{S} & 2.6\phantom{S} & 1.5\phantom{S} & \textbf{1.8}\phantom{S} \\

Baseline & & \multicolumn{1}{c||}{Primitives} & 259k\phantom{S} & \multicolumn{1}{c}{-} & 117k\phantom{S} & 311k\phantom{S} & 105k\phantom{S} & 431k\phantom{S} & 204k\phantom{S} & 514k\phantom{S} & 215k\phantom{S} & 254k\phantom{S} & 217k\phantom{S} & \textbf{263k}\phantom{S} \\

 & & \multicolumn{1}{c||}{Iterations} & 14.8k\phantom{S} & \multicolumn{1}{c}{-} & 13.7k\phantom{S} & 11.1k\phantom{S} & 3.8k\phantom{S} & 12.3k\phantom{S} & 8.8k\phantom{S} & 12.2k\phantom{S} & 12.4k\phantom{S} & 13.3k\phantom{S} & 14.8k\phantom{S} & \textbf{11.7k}\phantom{S} \\ \addlinespace[\topTableSpacing] \hline \addlinespace[\topTableSpacing]

 & & \multicolumn{1}{c||}{GPU (GiB)} & 6.8\phantom{S} & \multicolumn{1}{c}{-} & 5.9\phantom{S} & 6.7\phantom{S} & 5.7\phantom{S} & 6.8\phantom{S} & 6.2\phantom{S} & 7.4\phantom{S} & 6.3\phantom{S} & 6.8\phantom{S} & 6.4\phantom{S} & 6.5\phantom{S} \\

Optical Flow & & \multicolumn{1}{c||}{Primitives} & 371k\phantom{S} & \multicolumn{1}{c}{-} & 186k\phantom{S} & 452k\phantom{S} & 131k\phantom{S} & 444k\phantom{S} & 248k\phantom{S} & 629k\phantom{S} & 399k\phantom{S} & 357k\phantom{S} & 250k\phantom{S} & 347k\phantom{S} \\

 & & \multicolumn{1}{c||}{Iterations} & 13.7k\phantom{S} & \multicolumn{1}{c}{-} & 12.8k\phantom{S} & 9.1k\phantom{S} & 3.4k\phantom{S} & 11.6k\phantom{S} & 8.0k\phantom{S} & 10.8k\phantom{S} & 10.8k\phantom{S} & 11.6k\phantom{S} & 14.1k\phantom{S} & 10.6k\phantom{S} \\ \addlinespace[\topTableSpacing] \hline \addlinespace[\topTableSpacing]

 & & \multicolumn{1}{c||}{GPU (GiB)} & 1.8\phantom{S} & \multicolumn{1}{c}{-} & 1.3\phantom{S} & 1.9\phantom{S} & 1.4\phantom{S} & 1.8\phantom{S} & 1.7\phantom{S} & 2.3\phantom{S} & 1.8\phantom{S} & 1.8\phantom{S} & 2.1\phantom{S} & \textbf{1.8}\phantom{S} \\
 
MLP & & \multicolumn{1}{c||}{Primitives} & 412k\phantom{S} & \multicolumn{1}{c}{-} & 188k\phantom{S} & 399k\phantom{S} & 105k\phantom{S} & 475k\phantom{S} & 281k\phantom{S} & 577k\phantom{S} & 370k\phantom{S} & 329k\phantom{S} & 281k\phantom{S} & 342k\phantom{S} \\

 & & \multicolumn{1}{c||}{Iterations} & 14.2k\phantom{S} & \multicolumn{1}{c}{-} & 13.1k\phantom{S} & 10.6k\phantom{S} & 3.8k\phantom{S} & 12.3k\phantom{S} & 8.6k\phantom{S} & 11.5k\phantom{S} & 11.6k\phantom{S} & 12.8k\phantom{S} & 14.5k\phantom{S} & 11.3k\phantom{S} \\ \addlinespace[\topTableSpacing] \hline \addlinespace[\topTableSpacing]

 & & \multicolumn{1}{c||}{GPU (GiB)} & 7.6\phantom{S} & \multicolumn{1}{c}{-} & 6.1\phantom{S} & 7.8\phantom{S} & 5.7\phantom{S} & 8.2\phantom{S} & 6.5\phantom{S} & 7.4\phantom{S} & 6.3\phantom{S} & 7.0\phantom{S} & 7.3\phantom{S} & 7.0\phantom{S} \\
 
Optical Flow + & & \multicolumn{1}{c||}{Primitives} & 514k\phantom{S} & \multicolumn{1}{c}{-} & 281k\phantom{S} & 535k\phantom{S} & 131k\phantom{S} & 545k\phantom{S} & 368k\phantom{S} & 687k\phantom{S} & 342k\phantom{S} & 468k\phantom{S} & 367k\phantom{S} & 424k\phantom{S} \\

MLP & & \multicolumn{1}{c||}{Iterations} & 13.0k\phantom{S} & \multicolumn{1}{c}{-} & 11.8k\phantom{S} & 9.1k\phantom{S} & 3.4k\phantom{S} & 10.9k\phantom{S} & 7.6k\phantom{S} & 10.4k\phantom{S} & 11.4k\phantom{S} & 11.1k\phantom{S} & 13.5k\phantom{S} & 9.9k\phantom{S} \\ \addlinespace[\bottomTableSpacing] \bottomrule[1.5pt]

\end{tabular}
}

\vspace{-\baselineskip}
\label{table:results_qualityandperformancemetricsablations}

\end{table}

\subsection{\texttt{\upshape{Parking}} Sequence}

\begin{figure}[t]
    \centering
    \begin{center}
    \begin{tabular}{@{} c c c }
        \includegraphics[width=0.15\textwidth]{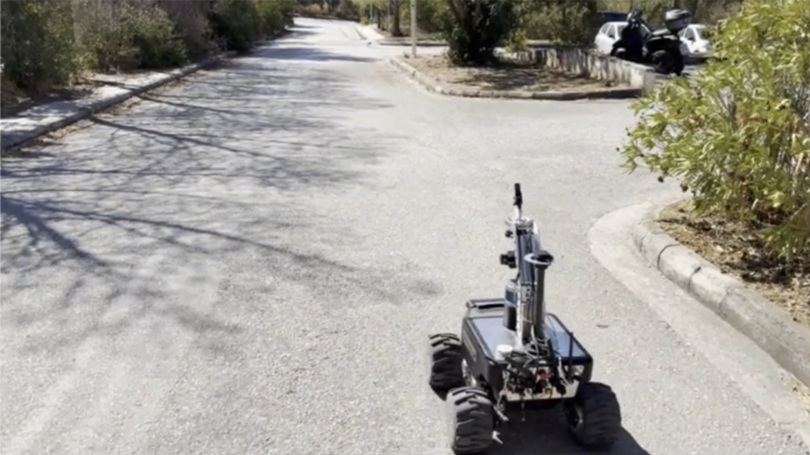} & \includegraphics[width=0.15\textwidth]{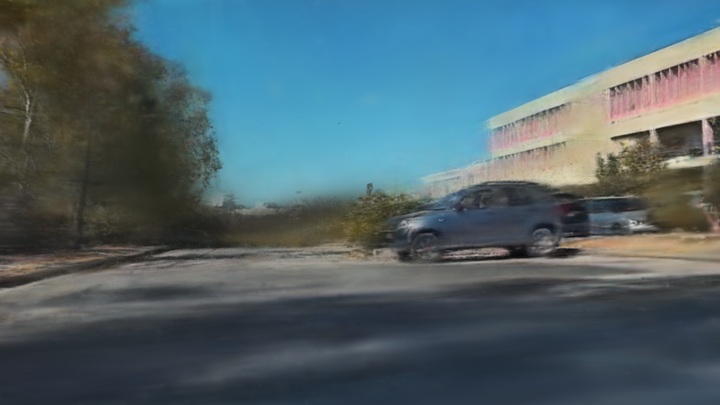} &
        \includegraphics[width=0.15\textwidth]{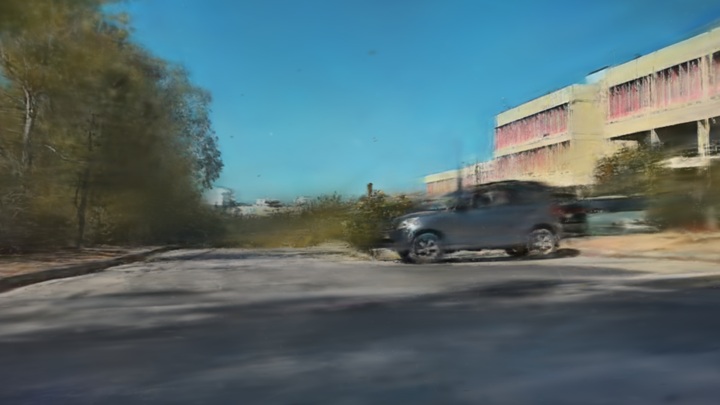} \\
        \scriptsize{(a) Ground vehicle} & \scriptsize{(b) Ours (baseline)} & \scriptsize{(c) Ours (OF+MLP) } \\ [0.1cm] 
        \includegraphics[width=0.15\textwidth]{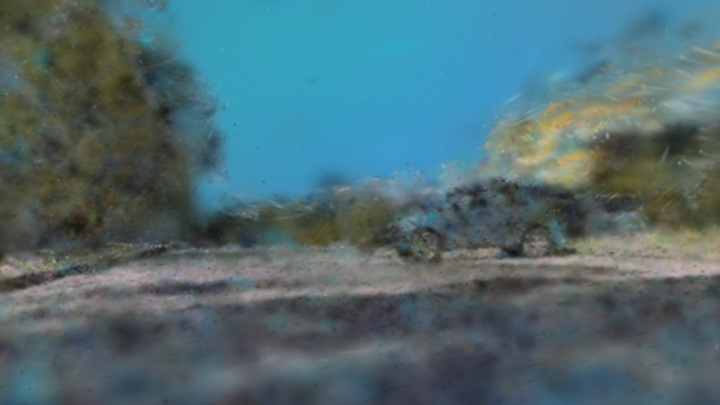} & \includegraphics[width=0.15\textwidth]{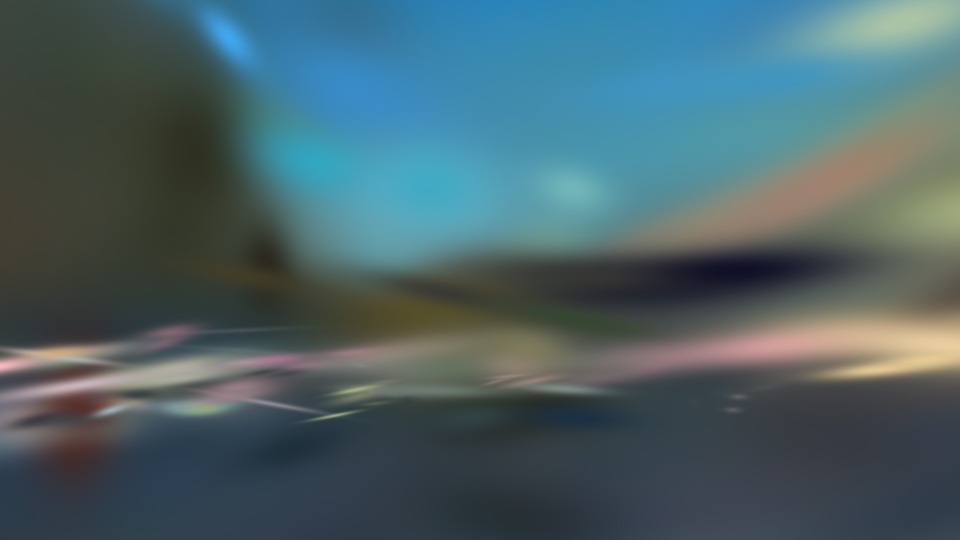} & \includegraphics[width=0.15\textwidth]{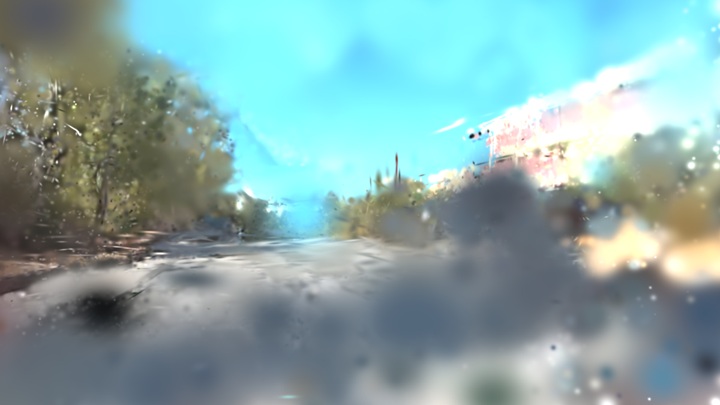} \\
        \scriptsize{(d) GigaSLAM} & \scriptsize{(e) PhotoSLAM} & \scriptsize{(f) MonoGS} \\ [0.1cm]
    \end{tabular}
    \vspace{-\baselineskip}
    \end{center}
    
    \vspace{-0.5\baselineskip}
    \caption{\textbf{Qualitative evaluation on the \texttt{Parking} sequence:} (a) The data acquisition UGV. (b)--(f) Comparative rendering results.}
    \label{fig:supplementary_groundvehicle_parking}
\end{figure}

To evaluate our contributions in a real-world setting, we deployed a ground vehicle robot equipped with a Viewpro Z10TIR RGB sensor to capture an outdoor sequence in an urban parking lot. Figure~\ref{fig:supplementary_groundvehicle_parking}(a) illustrates the platform and qualitative comparisons (b-f), while Table~\ref{table:supplementary_qualitymetricsextendedEMP} provides the quantitative photometric results, with both demonstrating significant improvements compared to all other state-of-the-art methods.
As indicated in Table~\ref{table:supplementary_qualitymetricsextendedEMP}, the spatial decomposition component yields only marginal gains due to minimal illumination and appearance change. Conversely, flow-densification provides a clearer improvement compared to the baseline, as seen in the details of Fig.~\ref{fig:supplementary_groundvehicle_parking}(b,c).

\section{Conclusion}
    
This paper presents a real-time monocular Gaussian Splatting SLAM system for large-scale operation, leveraging a robust, feature-based ORB-SLAM2 frontend to establish the geometric prior for our structured Gaussian anchor representation.
To bridge the gap between this sparse tracking frontend and the dense seeding demanded by Gaussian optimization, we formulate a flow-driven densification scheme that derives dense structural priors using epipolar geometry.
Furthermore, to handle environmental variability in long-term outdoor sequences, we introduce a scene partitioning strategy to efficiently model distinct local regions. 
Ablation studies validate the efficacy of our contributions, and a comprehensive evaluation on challenging outdoor datasets demonstrates the performance of our proposed system.
Our system advances state-of-the-art reconstruction quality while delivering real-time performance and scalable GPU memory usage, addressing the two major bottlenecks that limit the practical deployment of Gaussian Splatting SLAM systems.
 
\begin{table}[t]
\setlength{\tabcolsep}{7pt}

\settowidth{\wexp}{Static Line Trackletsssssssssssss}
\newcolumntype{C}{>{\raggedleft\arraybackslash}p{\dimexpr(\wexp-2\tabcolsep)/2}}

\settowidth{\xexp}{15.020}
\newcolumntype{D}{>{\centering\arraybackslash}p{\dimexpr(\xexp)}}

\settowidth{\yexp}{0.4100000}
\newcolumntype{E}{>{\centering\arraybackslash}p{\dimexpr(\yexp)}}

\centering
\caption{Photometric reconstruction quality results on the \texttt{Parking} sequence.}
\vspace{-0.75\baselineskip}
\resizebox{0.49\textwidth}{!}{%
\begin{tabular}{@{} C || *{7}{E} }
\toprule[1.5pt] \addlinespace[\topTableSpacing]
\multicolumn{1}{l||}{\textbf{Ablations} $\rightarrow$} & Baseline & OF & MLP &  OF+MLP & GigaSLAM & PhotoSLAM & MonoGS \\
  
\multicolumn{1}{l||}{\textit{Total frames}} & 2200 & 2200 & 2200 & 2200 & 2200 & 2200 & 2200 \\ \addlinespace[\topTableSpacing] \hline \addlinespace[\topTableSpacing]

PSNR $\uparrow$\phantom{i} & 23.428 & 23.648 & 23.459 & \textbf{23.719} & 19.149 & 21.066 & 13.351 \\

SSIM $\uparrow$\phantom{i} & \phantom{6}0.963 & \phantom{6}\textbf{0.964} & \phantom{6}0.963 & \phantom{6}\textbf{0.964} & \phantom{6}0.911 & \phantom{6}0.612 & \phantom{6}0.492 \\

LPIPS $\downarrow$\phantom{i} & \phantom{6}0.384 & \phantom{6}0.362 & \phantom{6}0.384 & \phantom{6}\textbf{0.359} & \phantom{6}0.719 & \phantom{6}0.659 & \phantom{6}0.750 \\ \addlinespace[\topTableSpacing] \bottomrule[1.5pt]

\end{tabular}
}
\vspace{-1\baselineskip}
\label{table:supplementary_qualitymetricsextendedEMP}
\end{table}

\bibliographystyle{IEEEtran}
{\hypersetup{hidelinks} \bibliography{mybibliography} }

\end{document}